\documentclass[12pt]{article}
\usepackage{times}
\usepackage{graphicx}
\usepackage{color}
\usepackage{multirow}
\usepackage{microtype}
\usepackage{booktabs} 
\usepackage{amsmath, amssymb, amsthm}
\usepackage{ascmac}
\usepackage[authoryear]{natbib}
\usepackage{rotating}
\usepackage{bbm}
\usepackage{latexsym}

\usepackage{hyperref}
\usepackage[subrefformat = parens]{subcaption}

\usepackage{mathtools}
\mathtoolsset{showonlyrefs, showmanualtags}
\newcommand{\bs}[1]{\boldsymbol{#1}}

\newcommand*{\tr}{\mathrm{tr}}
\newcommand*{\vect}{\mathrm{vec}}
\newcommand*{\diag}{\mathrm{diag}}
\newcommand*{\CN}{\mathcal{C}\mathcal{N}}

\newcommand{\re}{\mathrm{Re}}
\newcommand{\im}{\mathrm{Im}}

\newcommand{\captionfonts}{\normalsize}

\makeatletter  
\long\def\@makecaption#1#2{%
  \vskip\abovecaptionskip
  \sbox\@tempboxa{{\captionfonts #1: #2}}%
  \ifdim \wd\@tempboxa >\hsize
    {\captionfonts #1: #2\par}
  \else
    \hbox to\hsize{\hfil\box\@tempboxa\hfil}%
  \fi
  \vskip\belowcaptionskip}
\makeatother   

\begin{document}
\hspace{13.9cm}1

\ \vspace{20mm}\\

{\LARGE Gaussian Process Koopman Mode Decomposition}

\ \\
{\bf \large Takahiro Kawashima$^{\displaystyle 1}$, Hideitsu Hino$^{\displaystyle 1, \displaystyle 2, \displaystyle 3}$}\\
{$^{\displaystyle 1}$Department of Statistical Science, Graduate University of Advanced Studies (SOKENDAI), Tokyo, Japan.}\\
{$^{\displaystyle 2}$Institute of Statistical Mathematics, Tokyo, Japan.}\\
{$^{\displaystyle 3}$RIKEN AIP, Tokyo, Japan.}\\
%

\thispagestyle{empty}
\markboth{}{NC instructions}
\ \vspace{-0mm}\\
%
\begin{center} {\bf Abstract} \end{center}
In this paper, we propose a nonlinear probabilistic generative model of Koopman mode decomposition based on an unsupervised Gaussian process.
Existing data-driven methods for Koopman mode decomposition have focused on estimating the quantities specified by Koopman mode decomposition, namely, eigenvalues, eigenfunctions, and modes.
Our model enables the simultaneous estimation of these quantities and latent variables governed by an unknown dynamical system.
Furthermore, we introduce an efficient strategy to estimate the parameters of our model by low-rank approximations of covariance matrices.
Applying the proposed model to both synthetic data 
and a real-world epidemiological dataset,
we show that various analyses are available using the estimated parameters.

\section{Introduction}
\label{intro}
Many real-world phenomena are observed as multivariate (time) series data.
Although they sometimes appear to be disorderly, the obtained data may be governed by some intrinsic law.
Because such laws are expressed as dynamical systems in many fields,
the development of data-driven approaches to understand unknown dynamical systems is probably inevitable.\par

One data-driven strategy for dynamical systems is to employ with state space models, classically represented by
the Kalman filter \citep{kalman1960}, ensemble Kalman filter \citep{evensen2003},
particle filter \citep{gordon1993, kitagawa1996}, and 4D-Var \citep{lewis1985, dimet1986}.
An alternative approach is mode decomposition, which extracts some oscillating components from data.
If some background knowledge validates the assumption of a dynamical system,
we can comprehend data by estimating time-invariant parameters, including the modes.\par

\textit{Koopman mode decomposition} (KMD) enables us to specify the quantities to be estimated
on the basis of the \textit{Koopman operator theory} \citep{mezic2005, rowley2009}.
Although only limited special systems enable analytic calculations of the quantities,
\textit{dynamic mode decomposition} (DMD) provides a general data-driven algorithm to approximate them \citep{rowley2009, schmid2010}.
DMD is primitively divided into two types: the Arnoldi type \citep{rowley2009} and SVD-based type \citep{schmid2010}.
Both types give a simple linear approximation of the dynamics on an observation space,
thus various DMD extensions have been developed in the last decade \citep{jovanovic2014, dawson2016, leclainche2017, heas2020}.
To overcome the limitations of linear approximations,
some nonlinear extensions of DMD have been proposed on the basis of user-defined bases \citep{williams2015},
kernel methods \citep{williams2015a, kawahara2016}, or neural networks \citep{takeishi2017a}.
Nonetheless, nonlinear probabilistic generative models of KMD have not yet been studied, as mentioned in Section \ref{works}.

In this study, we develop a nonlinear generative model for KMD with an unsupervised Gaussian process (GP)
named \textit{Gaussian process Koopman mode decomposition} (GPKMD).
An existing unsupervised GP method for dynamical systems known as
\textit{Gaussian process dynamical model} (GPDM) \citep{wang2005} already exists.
The GPDM was derived from the \textit{Gaussian process latent variable model} (GPLVM) \citep{lawrence2005},
which is the GP form of \textit{probabilistic principal component analysis} (probabilistic PCA),
and can be viewed as a GP-based extension of an autoregressive model.
Whereas GPLVM and GPDM only focus on dimensionality reduction or learning nonlinear mappings from a latent space to an observation space,
our method can be used to estimate the latent variables and quantities of KMD simultaneously.\par

This paper has the following main contributions:
\begin{itemize}
    \item We provide a novel perspective of KMD through the GP-based nonlinear generative model named GPKMD.
          The generative modeling enables us to estimate not only the quantities specified by KMD but also the latent variables 
          and enables us to obtain richer information from estimands.
    \item We propose an efficient computing strategy for GPKMD using low-rank approximations of Gram matrices and matrix diagonalization. We show that the complexity of our strategy is markedly superior to the existing one.
    \item We demonstrate our proposed method on synthetic data generated from a nonlinear limit cycle and a real-world epidemiological dataset. We show the usefulness of the proposed method for interpreting the data from various viewpoints.
\end{itemize}

\subsection{Koopman Mode Decomposition}
\textit{Koopman mode decomposition} (KMD) is a framework to transform multidimensional series data into
a tractable sum-of-modes representation.
We provide a brief introduction to KMD.\par
Let the latent variables $\bs{x}_t \in \mathcal{X} \subset \mathbb{R}^P$ be evolved deterministically by an unknown map
$\bs{f}: \mathcal{X} \to \mathcal{X}$,
\begin{align}
    \label{eq:dynamics}
    \bs{x}_{t+1} = \bs{f}(\bs{x}_{t}).
\end{align}
Observations that we can treat are obtained through an observable $\mathcal{G} \ni g: \mathcal{X} \to \mathbb{C}$ as $g(\bs{x}_t)$,
where $\mathcal{G}$ is an appropriate complex-valued function space.
The \textit{Koopman operator} $\mathcal{K}: \mathcal{G} \to \mathcal{G}$ is defined as an operator
that maps the observable at $t$ to that at $t+1$:
\begin{align}
    (\mathcal{K} g)(\bs{x}_t) = (g \circ \bs{f})(\bs{x}_{t}) = g(\bs{x}_{t+1}).
\end{align}
Although we considered the latent dynamics $\bs{f}$ above,
the Koopman operator $\mathcal{K}$ can also describe the evolution of a system on the function space $\mathcal{G}$.
Despite the nonlinearity of $\bs{f}$,
the Koopman operator is linear owing to its lifting to the infinite-dimensional space.
This property permits the spectral decomposition of $\mathcal{K}$,
\begin{align}
    \label{eq:spectrum}
     \mathcal{K} \phi_k = \lambda_k \phi_k,
\end{align}
where $\lambda_k \in \mathbb{C}$ and $\phi_k: \mathcal{X} \to \mathbb{C}$ are the $k$-th
\textit {Koopman eigenvalue} and the corresponding \textit{Koopman eigenfunction}, respectively.
Suppose there are $D$ distinct observables $g_1,\dots,g_D$ such that $g_d \in \mathcal{G},d=1,\dots,D$,
then we define a $D$-dimensional observation 
$\bs{y}_t = \bs{g}(\bs{x}_t) = (g_1(\bs{x}_t), \ldots, g_D(\bs{x}_t))^\top \in \mathbb{C}^D$.
Assuming the $D$-dimensional observable $\bs{g}$ is expanded by Koopman eigenfunctions $\{\phi_k\}$,
we obtain
\begin{align}
    \label{eq:expansion}
    \bs{y}_t = \bs{g}(\bs{x}_t) = \sum^\infty_{k = 1} \phi_k(\bs{x}_t) \bs{w}_k,
\end{align}
where $\bs{w}_k \in \mathbb{C}^D$ is the $k$-th coefficient called the \textit{Koopman mode}.
By applying spectral decomposition \eqref{eq:spectrum} to \eqref{eq:expansion},
observations can be transformed recurrently as
\begin{align}
    \label{eq:kmd}
    \bs{y}_t = \bs{g}(\bs{x}_{t}) = (\mathcal{K} \bs{g})(\bs{x}_{t-1})
    &= \sum^\infty_{k = 1} (\mathcal{K} \phi_k)(\bs{x}_{t - 1}) \bs{w}_k \nonumber\\
                        &= \sum^\infty_{k = 1} \lambda_k \phi_k(\bs{x}_{t - 1}) \bs{w}_k \nonumber\\
                        &= \cdots = \sum^\infty_{k = 1} \lambda^{t}_k \phi_k(\bs{x}_0) \bs{w}_k.
\end{align}
Given observations $\bs{y}_1, \ldots, \bs{y}_T$,
we can unravel the hidden dynamics governing the system by estimating the Koopman quantities
$\{\lambda_k\}, \{\phi_k\}$, and $\{\bs{w}_k\}$ in \eqref{eq:kmd}, instead of $\bs{f}$.
KMD is the framework used to understand the data-generating system with this scheme.
The inferable quantities depend on the algorithm;
for example, DMD approximates $\bs{f}$ by low-rank linear dynamics and provides the finite sets $\{\lambda_k\}$ and $\{\bs{w}_k\}$.

\subsection{Organization}
The rest of the paper is organized as follows.
In Section \ref{works}, we briefly review related studies with focus on Bayesian perspectives of KMD or DMD.
Our proposed method, GPKMD, is introduced in Section \ref{gpkmd}.
In Section \ref{scale}, we also propose a low-rank approximation method for the GPKMD likelihood towards scalable inference.
We demonstrate our method on both synthetic and real-world data in Section \ref{ex}.
In Section \ref{discussion}, we discuss some key points regarding this study, including limitations and future works,
and we present our conclusion.

\section{Related Works}
\label{works}
\subsection{Gaussian Processes and KMD}
In previous works, researchers have attempted to connect KMD and GP regressions.
\citet{masuda2019} proposed a GP-based algorithm for Arnoldi-type DMD.
This algorithm determines coefficients of the companion matrix based on the prediction by GP regression,
which is conditioned by past observations.
Although the method employs GP regression,
it requires \textit{a posteriori} deterministic matrix factorization processes to obtain the Koopman eigenvalues and modes.
Therefore, the advantages of probabilistic methods and interpretability are limited.
\citet{lian2020} studied a model predictive control method based on Koopman operator theory.
Because the work focused on the control design, they did not discuss an inference framework for Koopman quantities.\par

Estimating Koopman quantities can be regarded as an inverse problem.
This perspective implies that KMD is essentially an unsupervised task;
therefore, as a complementary to the existing works, we take an unsupervised approach.

\subsection{Bayesian Models of KMD}
In some Bayesian models, KMD (or DMD) is treated as unsupervised learning.
\citet{takeishi2017} proposed \textit{Bayesian DMD} and an efficient sampling algorithm for the posterior.
In the Bayesian DMD model, each output of the Koopman eigenfunction $\phi_k(\bs{x}_t)$ in \eqref{eq:kmd} is parameterized as a scalar-valued i.i.d. random variable $\phi_{kt}$.
However, this simplification may discard important structures in the eigenfunctions $\{\phi_k\}$ and latent variables $\{\bs{x}_t\}$. 
To alleviate this shortcoming, the \textit{Bayesian DMD with variational matrix factorization} (BDMD-VMF) model was developed \citep{kawashima2021}.
By rewriting \eqref{eq:kmd} as $\bs{y}_t \approx \sum_k \lambda^{t}_k \bs{w}_k \phi_{k0}$,
BDMD-VMF avoids the explicit treatment of the eigenfunctions.
Moreover, BDMD-VMF employs VMF \citep{lim2007, nakajima2011} to determine its prior and marginalize its higher-dimensional parameters;
thus, the computational stability is improved even for incomplete observations.\par

In this study, we develop a GP-based generative model of KMD as an extension of Bayesian DMD.
Whereas both Bayesian DMD and BDMD-VMF are based on linear parameterizations of the output of the Koopman eigenfunctions, not the latent variables $\{\bs{x}_t\}$,
our model explicitly incorporates the latent variables as model parameters (i.e., random variables).
To the best of our knowledge, this is the first work enabling the latent variables to be directly estimated from observations in the framework of KMD.

\section{Gaussian Process Koopman Mode Decomposition}
\label{gpkmd}
Gaussian processes (GPs) are representative nonparametric methods of learning nonlinear mappings from an input space $\mathcal{X}$ to an output space $\mathcal{Y}$.
By formulating KMD as an unsupervised GP, we establish a nonlinear generative model of KMD.\par

Let $\bs{Y} = (\bs{y}_1, \ldots, \bs{y}_T) \in \mathbb{C}^{D \times T}$ be the data matrix and
$\bs{X} = (\bs{x}_0, \ldots, \bs{x}_T) \in \mathbb{R}^{P \times (T+1)}$ be the latent variables.
We start by assuming that the value of each Koopman eigenfunction $\phi_k$ evaluated as any $\bs{x} \in \mathbb{R}^{P}$ is represented as the inner product $\langle \cdot, \cdot \rangle_\mathcal{H}$
on a reproducing kernel Hilbert space (RKHS) $\mathcal{H}$.
We then expand as
\begin{align}
    \label{eq:eigfunc_expanded}
    \phi_k(\bs{x}) = \langle \bs{b}_k, \bs{\psi}(\bs{x}) \rangle_{\mathcal{H}} = \sum_l b_{kl} \psi_l(\bs{x})
\end{align}
using coefficients $\bs{b}_k = (b_{k1}, b_{k2}, \ldots) \in \mathcal{H}$ and the feature map $\bs{\psi} = (\psi_1, \psi_2, \ldots) : \mathcal{X} \to \mathcal{H}$.
We define the likelihood of KMD by incorporating the equalities \eqref{eq:expansion} and \eqref{eq:kmd} up to first-order,
\begin{align}
    p(\bs{y}_{t} | \{\bs{x}_t\}, \{\lambda_k\}, \{\bs{w}_k\}, \{b_{kl}\}, \sigma^2) 
    &= \CN \left ( \bs{y}_{t} \left | \sum^K_{k = 1} \left ( \sum_l b_{kl} \psi_l(\bs{x}_t) \right ) \bs{w}_k, \sigma^2 \bs{I} \right . \right ) \\
    \label{eq:pre_lik}
    &\times \CN \left ( \bs{y}_{t} \left | \sum^K_{k = 1} \lambda_k \left ( \sum_l b_{kl} \psi_l(\bs{x}_{t-1}) \right ) \bs{w}_k, \sigma^2 \bs{I} \right . \right ),
\end{align}
where $\CN(\cdot, \cdot)$ denotes a complex normal distribution.
In \eqref{eq:pre_lik}, the countably infinite summations of modes are truncated at $K$.
The expansion coefficients $\{b_{kl}\}$ can be marginalized out from each $\CN(\cdot, \cdot)$
in the likelihood \eqref{eq:pre_lik}:
with the i.i.d. prior $p(b_{kl}) = \CN(b_{kl} | 0, \sigma^2_b/2) \propto \CN(b_{kl} | 0, \sigma^2_b)^2$.
We then obtain the following marginalized likelihood (see Appendix \ref{appendix} for derivation details):
\begin{align}
    p(\bs{Y} | \bs{X}, \bs{\Lambda}, \bs{W}, \sigma^2, \sigma^2_b)
    &= \CN(\vect(\bs{Y}) | \bs{0}, \sigma^2 \bs{I} + \sigma^2_b (\bs{K}_{1} \otimes \bs{W}\bs{W}^\ast))\\
    \label{eq:lik}
    &\times \CN(\vect(\bs{Y}) | \bs{0}, \sigma^2 \bs{I} + \sigma^2_b (\bs{K}_{0} \otimes \bs{W}\bs{\Lambda}\bs{\Lambda}^\ast\bs{W}^\ast)),~~
\end{align}
where $\bs{W} = (\bs{w}_1, \ldots, \bs{w}_K)$ and $\bs{\Lambda} = \diag(\{\lambda_k\}^K_{k=1})$. 
$\bs{K}_1$ and $\bs{K}_0$ are $T \times T$ Gram matrices consisting of $\{\bs{x}_t\}^{T}_{t = 1}$ and $\{\bs{x}_t\}^{T-1}_{t = 0}$ with
a positive definite kernel $k(\bs{x}, \bs{x}') = \langle \bs{\psi}(\bs{x}), \bs{\psi}(\bs{x}') \rangle_{\mathcal{H}}$, respectively.
$\bs{W}^\ast$ denotes the Hermitian transpose of $\bs{W}$.
The marginalized likelihood \eqref{eq:lik} appears unnatural because it is divided into two terms, but
we can merge them into a single zero-mean $\CN(\cdot, \cdot)$.
Since the covariance matrices of the joint likelihood \eqref{eq:lik} are formed by the Gram matrices of latent variables,
we obtain the GP formulation for KMD.
We define \eqref{eq:lik} as the likelihood of our proposal, \textit{Gaussian process Koopman mode decomposition} (GPKMD).

\subsection{Prior Distributions for GPKMD}
We should also consider rational priors
for the parameters $\bs{X}, \bs{W}, \bs{\Lambda}, \sigma^2,$ and $\sigma^2_b$.
Similar to the configuration of KMD \eqref{eq:dynamics}, GPKMD should incorporate
latent dynamics explicitly as its prior in a probabilistic sense.
Thus, we adopt a GPDM-inspired prior for the latent variable $\bs{X}$ \citep{wang2005}.
That is, denoting $\bs{X}_1 = (\bs{x}_1, \ldots, \bs{x}_T) \in \mathbb{R}^{P \times T}$ and
a Gram matrix consisting of $\{\bs{x}_t\}^{T-1}_{t = 0}$ with a kernel function $k_x(\cdot, \cdot)$ by
$\bs{K}_X$,
we use
\begin{align}
    \label{eq:gpdm_prior}
    p(\bs{X}) = \mathcal{N}(\bs{x}_0 | \bs{0}, s^2_x \bs{I}) \mathcal{M}\mathcal{N}(\bs{X}_1 | \bs{O}, \bs{I}, \bs{K}_X + s^2_x \bs{I})
\end{align}
for the prior $p(\bs{X})$.
Here, $\mathcal{M}\mathcal{N}(\cdot, \cdot, \cdot)$ denotes a matrix normal distribution.
Note that the prior can be regarded as a GP extension of the first-order autoregressive model.
Unless there is a particular reason,
it is reasonable to employ simple priors for other parameters, such as
\begin{align}
    p(w_{dk}) &= \CN(w_{dk} | 0, s^2_w),\\
    p(\lambda_{k}) &= \CN(\lambda_{k} | 0, s^2_\lambda),\\
    p(\sigma^2) &= \mathrm{InvGamma}(\sigma^2 | \alpha, \beta),\\
    p(\sigma^2_b) &= \mathrm{InvGamma}(\sigma^2_b | \alpha_b, \beta_b).
\end{align}

\section{Scalable Inference}
\label{scale}
In theory, the posterior or its point estimates of GPKMD parameters
can be obtained using the marginal likelihood \eqref{eq:lik} with appropriate priors.
However, the GPKMD likelihood contains very large $DT \times DT$-sized covariance matrices, which inhibit scalable inference.
Straightforward computations of the GPKMD likelihood \eqref{eq:lik} and its gradients require
an extremely high computational cost of $\mathcal{O}(D^3T^3)$.
Hereafter, we tackle the scalability of GPKMD.
We only consider the first $\CN(\cdot, \cdot)$ in \eqref{eq:lik} for simplicity in this section, but the same approach applies to 
the second $\CN(\cdot, \cdot)$.
\par

\subsection{Stegle's Method}
Multioutput or multitask GPs often have Kronecker-structured covariance matrices, and
they are sometimes called \textit{Kronecker GPs} \citep{stegle2011}.
GPKMD can be considered a type of Kronecker GP.
\citet{stegle2011} and \citet{rakitsch2013} proposed an efficient inference method for
Kronecker GPs using an eigendecomposition-based trick.
First, consider the eigendecomposition
$\bs{K}_1 = \bs{U}_{K} \bs{S}_{K} \bs{U}^\top_{K}, ~\bs{W}\bs{W}^\ast = \bs{U}_{W} \bs{S}_{W} \bs{U}^\ast_{W}$.
Following Stegle's method, the inversion of the GPKMD covariance matrix is exactly transformed into
\begin{align}
    &(\sigma^2 \bs{I} + \sigma^2_b (\bs{K}_{1} \otimes \bs{W}\bs{W}^\ast))^{-1} \\
    \label{eq:stegle}
    &= (\underbrace{\bs{U}_{K} \otimes \bs{U}_{W}}_{DT \times DT})
    (\underbrace{\sigma^2 \bs{I} + \sigma^2_b(\bs{S}_{K} \otimes \bs{S}_{W})}_{DT \times DT ~ \mbox{\footnotesize (diagonal)}})^{-1}
    (\underbrace{\bs{U}_{K} \otimes \bs{U}_{W}}_{DT \times DT})^\ast.
\end{align}
Because the matrix to be inverted is reformed into a diagonal matrix, the complexity of the inversion is reduced to $\mathcal{O}(D^3 + T^3)$,
which is dominated by the eigendecomposition for $\bs{K}_1$ and $\bs{W}\bs{W}^\ast$.
$\log\det(\cdot)$ is similarly computed as
\begin{align}
    \log\det (\sigma^2 \bs{I} + \sigma^2_b (\bs{K}_{1} \otimes \bs{W}\bs{W}^\ast))
    = \log\det(\underbrace{\sigma^2 \bs{I} + \sigma^2_b(\bs{S}_{K} \otimes \bs{S}_{W})}_{DT \times DT ~ \mbox{\footnotesize (diagonal)}}),
\end{align}
and the gradients of the likelihood can also be converted to reduced forms.
\par

Stegle's method is effective for GPKMD; however, we still have some considerations:
\begin{itemize}
    \item For the interpretability, we often use a small number of Koopman modes, $K$, typically about 5--30.
    For $K \ll D$, the diagonal elements of the eigenvalue matrix $\bs{S}_W$ are sparse since $\mathrm{rank}(\bs{W}\bs{W}^\ast) = K$.
    This implies the possibility of further reducing in the computational cost.
    \item The space complexity of Stegle's method is $\mathcal{O}(D^2 + T^2)$. For large $D$ or/and $T$ (e.g., $>100,000$),
    ordinary computers may run out of memory.
\end{itemize}

\subsection{Low-rank Approximations}
In kernel methods, Gram matrices can be well approximated by low-rank matrices in many practical cases.
\citet{bonilla2007} proposed an efficient prediction strategy for multitask GPs by applying low-rank approximations to Gram matrices.
We propose a considerably more efficient strategy for various computations of GPKMD
by combining the above-explained Stegle's method and low-rank approximations.\par

By applying an appropriate algorithm (e.g., incomplete Cholesky decomposition or the Nystr\"om method~\citep{drineas2005}),
we can approximate the Gram matrix as $\bs{K}_1 \approx \bs{R}\bs{R}^\top$,
where $\bs{R} \in \mathbb{R}^{T \times S}$ for $S < T$.
If the Nystr\"om method is employed, we can obtain $\bs{C} \in \mathbb{R}^{T \times S}$ and $\bs{\Omega} \in \mathbb{R}^{S \times S}$
such that $\bs{K}_1 \approx \bs{C} \bs{\Omega} \bs{C}^\top$.
Then, the eigendecomposition of the symmetric matrix $\bs{\Omega}$ enables us to obtain the desired matrix $\bs{R}$.
Afterwards, by using thin SVD $\bs{R} = \bs{U}_{K} \bs{\Sigma}_{K} \bs{V}^T_{K},~ \bs{W} = \bs{U}_W \bs{\Sigma}_W \bs{V}^\ast_W$
and the Woodbury identity, the inverse covariance matrix of GPKMD is approximated as
\begin{align}
    (\sigma^2 \bs{I} + \sigma^2_b (\bs{K}_1 \otimes \bs{W}\bs{W}^\ast))^{-1}
    &\approx (\sigma^2 \bs{I} + \sigma^2_b(\bs{R} \bs{R}^\top \otimes \bs{W}\bs{W}^\ast))^{-1} \\
    &= \sigma^{-2} \bs{I} - \sigma^{-2} \sigma^2_b \underbrace{(\bs{U}_{K} \bs{\Sigma}_{K} \otimes \bs{U}_W \bs{\Sigma}_W)}_{DT \times KS}\\
    & \times (\underbrace{\sigma^2 \bs{I} + \sigma^2_b ( \bs{\Sigma}^2_{K} \otimes \bs{\Sigma}^2_W )}_{KS \times KS ~ \mbox{\footnotesize (diagonal)}})^{-1}
    \underbrace{(\bs{U}_{K} \bs{\Sigma}_{K} \otimes \bs{U}_W \bs{\Sigma}_W)^\ast}_{KS \times DT}.
\end{align}
On the other hand, the $\log\det(\cdot)$ of the covariance matrix can be transformed by the Weinstein--Aronszajn identity \citep{kato1995},
\begin{align}
    &\log\det(\sigma^2 \bs{I} + \sigma^2_b (\bs{K}_1 \otimes \bs{W}\bs{W}^\ast))\\
    &\approx \log\det(\sigma^2 \bs{I} + \sigma^2_b(\bs{R} \bs{R}^\top \otimes \bs{W}\bs{W}^\ast)) \\
    &=(DT - KS)\log\sigma^2
    +\log\det\underbrace{(\sigma^2 \bs{I} + \sigma^2_b (\bs{\Sigma}^2_{K} \otimes \bs{\Sigma}^2_{W}))}_{KS \times KS ~\mbox{\footnotesize (diagonal)}}.
\end{align}
Since the computational complexity of our approach is dominated by
the Nystr\"om method (or incomplete Cholesky decomposition) and SVD, it is markedly reduced to $\mathcal{O}(DK^2 + TS^2)$
for $K \ll D$ and $S \ll T$.
Notably, it is unnecessary to store the $T \times T$ matrix $\bs{K}_1$ (and $\bs{K}_0$) in the memory
in both the Nystr\"om and incomplete Cholesky decomposition algorithms.
Therefore, the space complexity of GPKMD can be reduced to $\mathcal{O}(DK + TS)$.
The gradients of GPKMD can also be evaluated in a short time, as shown in Appendix \ref{appendix}.

\section{Experiments}
\label{ex}
In this section, we demonstrate GPKMD for two experimental settings,
one with a synthetic dataset and one with a real-world dataset.
Through the experiments below, we show that a wide range of information about given data is available from the estimated parameters of GPKMD.
We employed MAP estimation by the conjugate gradient method for learning GPKMD.
The estimation of GPKMD parameters is sensitive to the initial values since
the posterior defined with \eqref{eq:lik} and \eqref{eq:gpdm_prior} is non-convex.
For the initial values of GPKMD, we used PCA results for the latent variables $\bs{X}$ and
standard DMD results for the Koopman eigenvalues $\{\lambda_k\}$ and modes $\{\bs{w}_k\}.$
For the kernel functions of GPKMD, we employed an RBF kernel for $k(\cdot, \cdot)$ in \eqref{eq:lik}
and an RBF+linear kernel for $k_x(\cdot, \cdot)$ in \eqref{eq:gpdm_prior}.

\subsection{Stuart--Landau Equation}
\begin{figure}[t]
    \centering
    \begin{minipage}{0.32\columnwidth}
        \centering
        \includegraphics[width = 1.0\columnwidth]{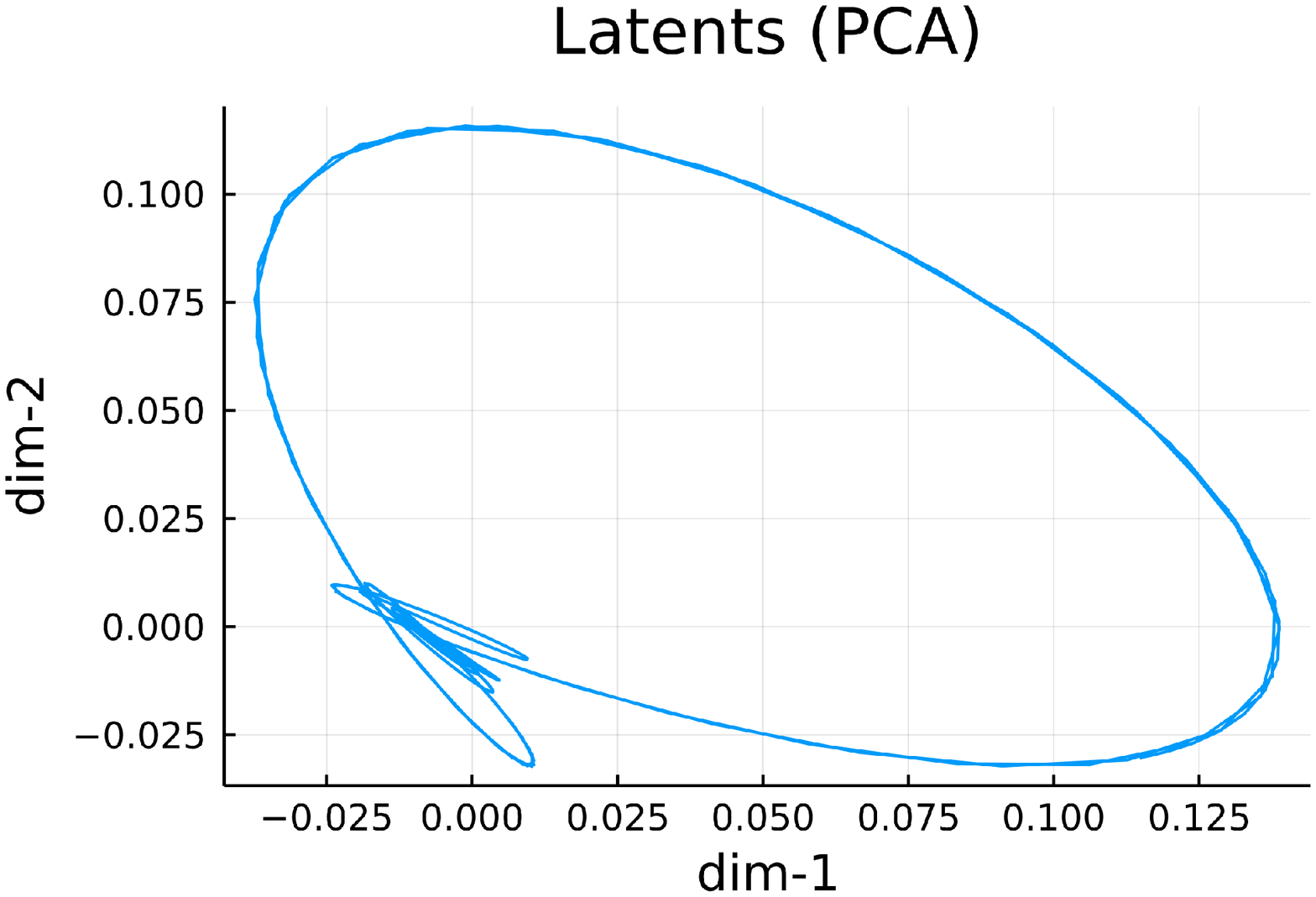}
        \subcaption{$\sigma_y = 0$, PCA}
        \label{subfig:sl0_trajectory_pca}
    \end{minipage}
    \begin{minipage}{0.32\columnwidth}
        \centering
        \includegraphics[width = 1.0\columnwidth]{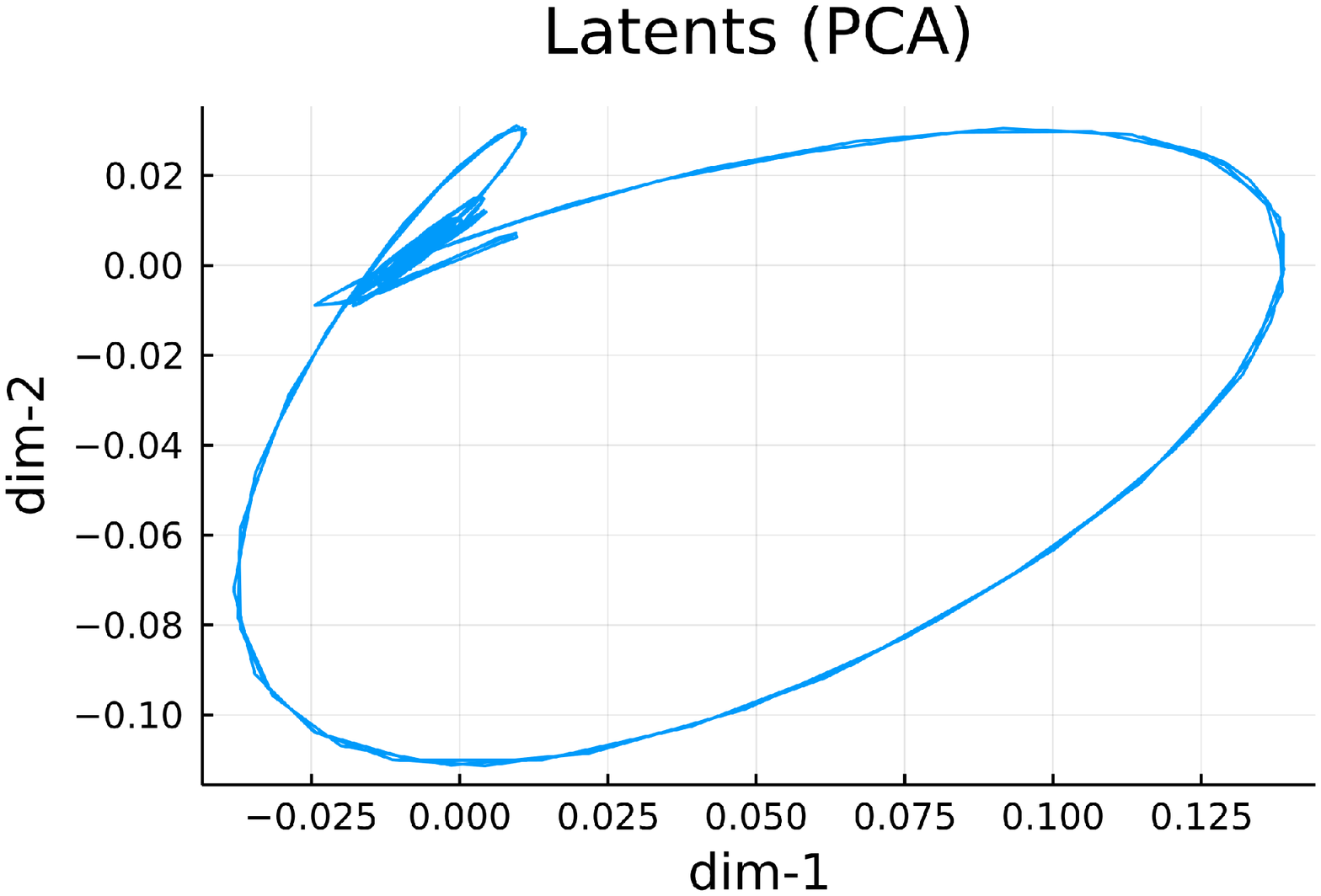}
        \subcaption{$\sigma_y = 0.01$, PCA}
        \label{subfig:sl001_trajectory_pca}
    \end{minipage}
    \begin{minipage}{0.32\columnwidth}
        \centering
        \includegraphics[width = 1.0\columnwidth]{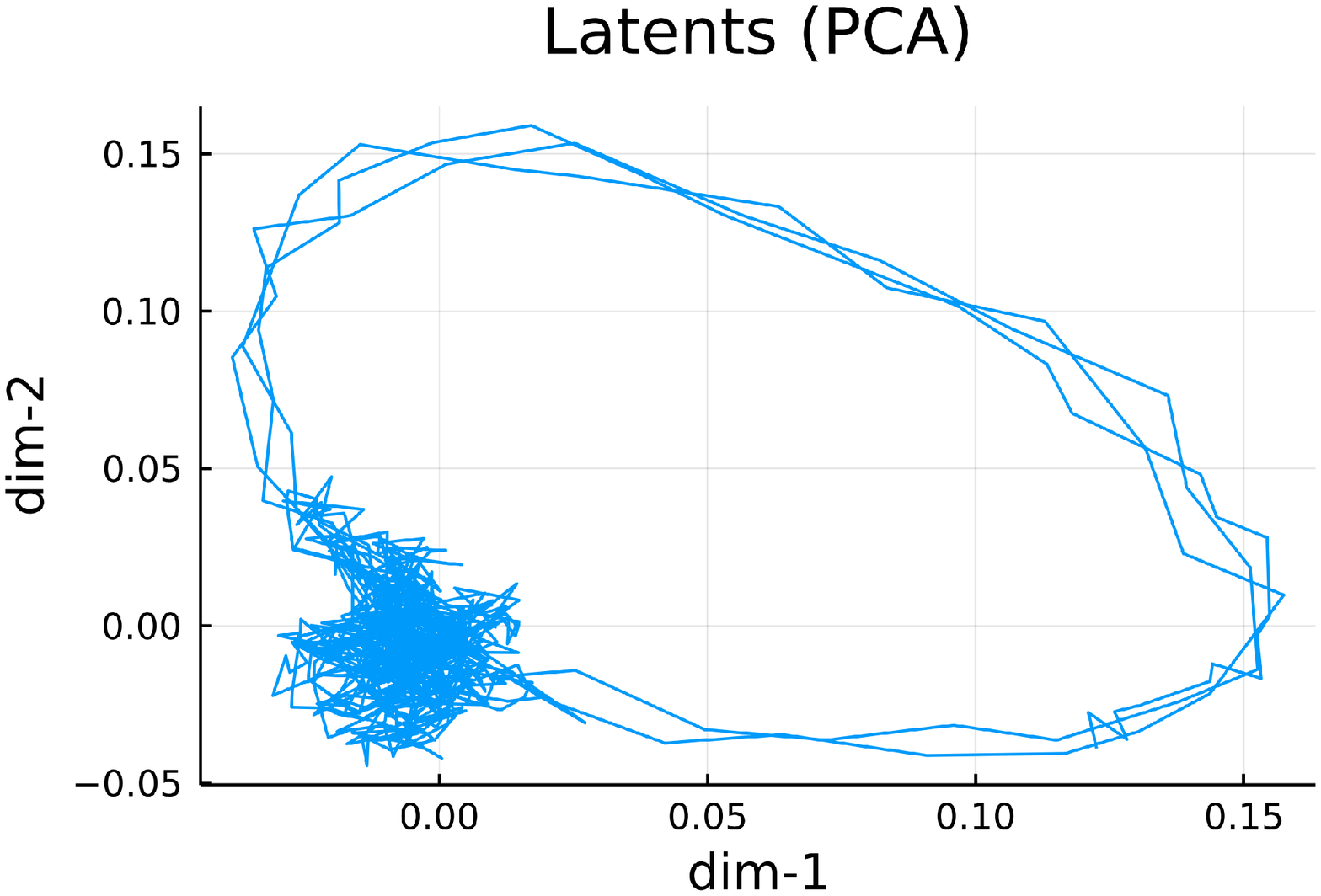}
        \subcaption{$\sigma_y = 0.2$, PCA}
        \label{subfig:sl02_trajectory_pca}
    \end{minipage}
    \\
    \begin{minipage}{0.32\columnwidth}
        \centering
        \includegraphics[width = 1.0\columnwidth]{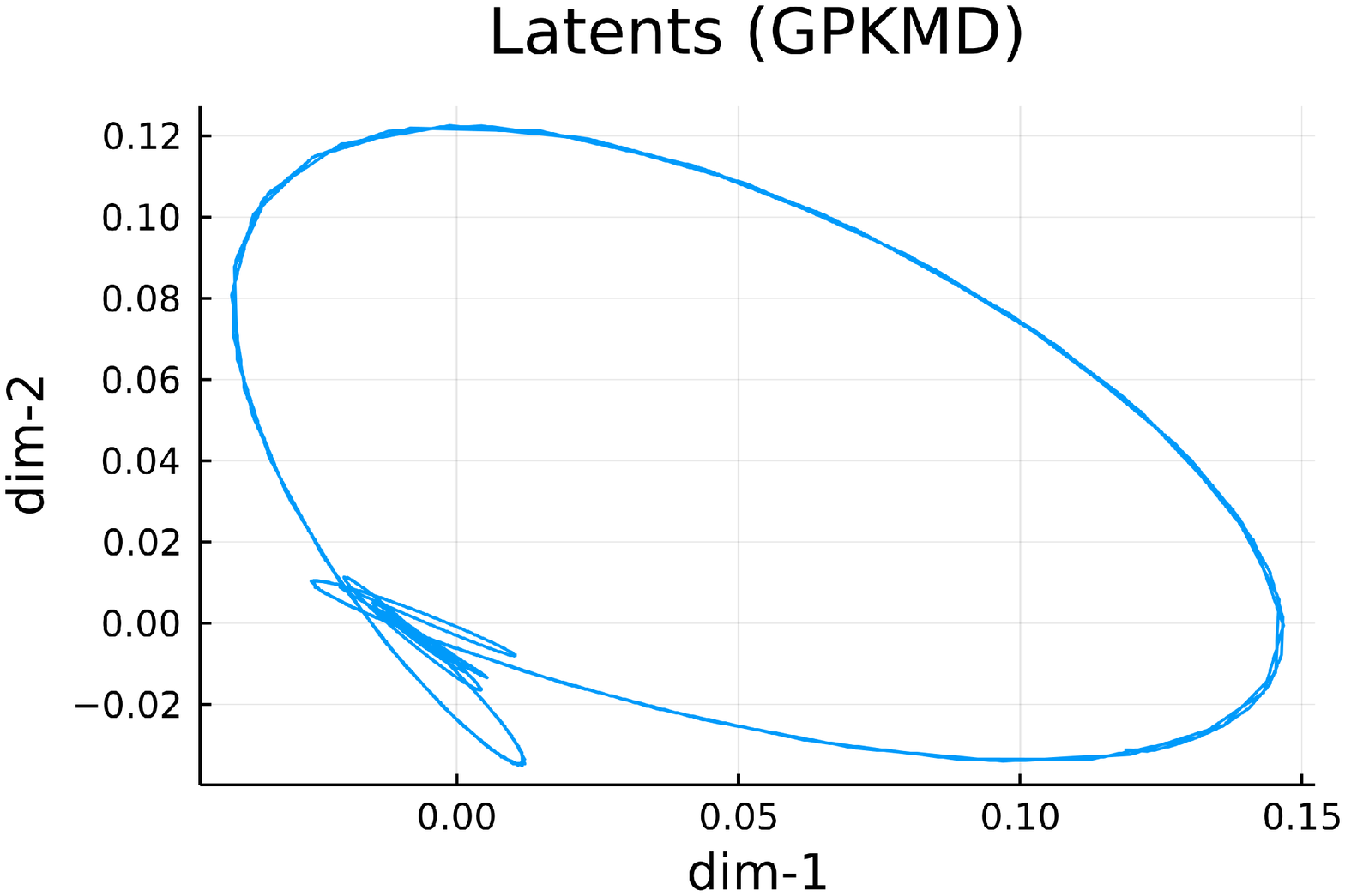}
        \subcaption{$\sigma_y = 0$, GPKMD}
        \label{subfig:sl0_trajectory_gpkmd}
    \end{minipage}
    \begin{minipage}{0.32\columnwidth}
        \centering
        \includegraphics[width = 1.0\columnwidth]{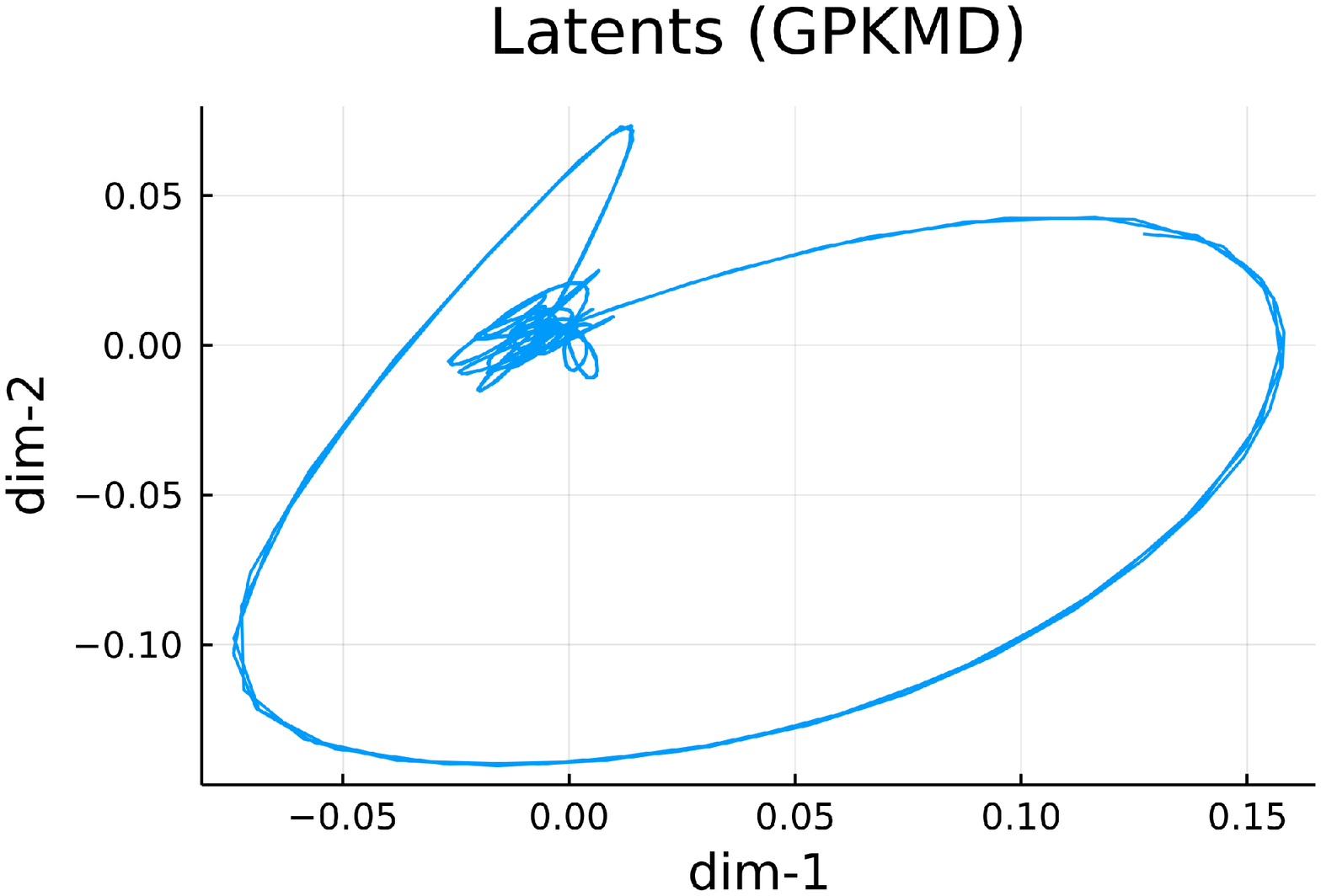}
        \subcaption{$\sigma_y = 0.01$, GPKMD}
        \label{subfig:sl001_trajectory_gpkmd}
    \end{minipage}
    \begin{minipage}{0.32\columnwidth}
        \centering
        \includegraphics[width = 1.0\columnwidth]{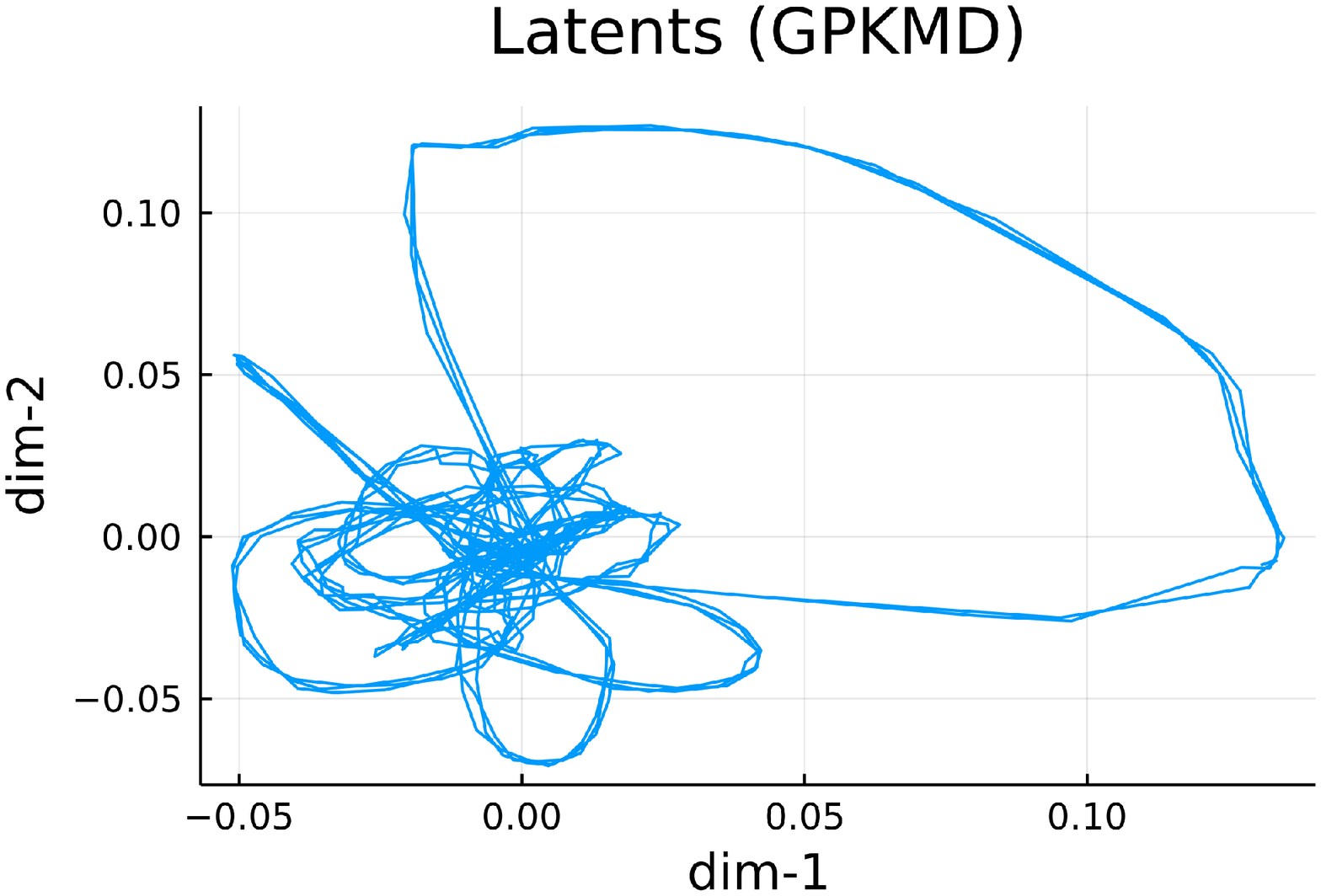}
        \subcaption{$\sigma_y = 0.2$, GPKMD}
        \label{subfig:sl02_trajectory_gpkmd}
    \end{minipage}
    \caption{
        Latent variables estimated by PCA and GPKMD for $P = 2$ and different noise levels,
        $\sigma_y = 0, 0.01, 0.2.$
    }
    \label{fig:sl_latents}
\end{figure}

First, we applied GPKMD to a synthetic dataset that follows the \textit{Stuart--Landau equation}.
The Stuart--Landau equation is a well-known nonlinear dynamical system
that has the discretized form
\begin{align}
    r_{t+1} &= r_t + (\delta r_t - r^3_t)\Delta t\\
    \theta_{t+1} &= \theta_t + (\gamma - \beta r^2_t) \Delta t
\end{align}
in polar coordinates.
The behavior of the system is determined by the parameters $\delta, \gamma,$ and $\beta$.
For example, $\delta > 0$ induces the limit cycle.\par
   
\begin{figure}[t]
    \centering
    \begin{minipage}{0.32\columnwidth}
        \centering
        \includegraphics[width = 1.0\columnwidth]{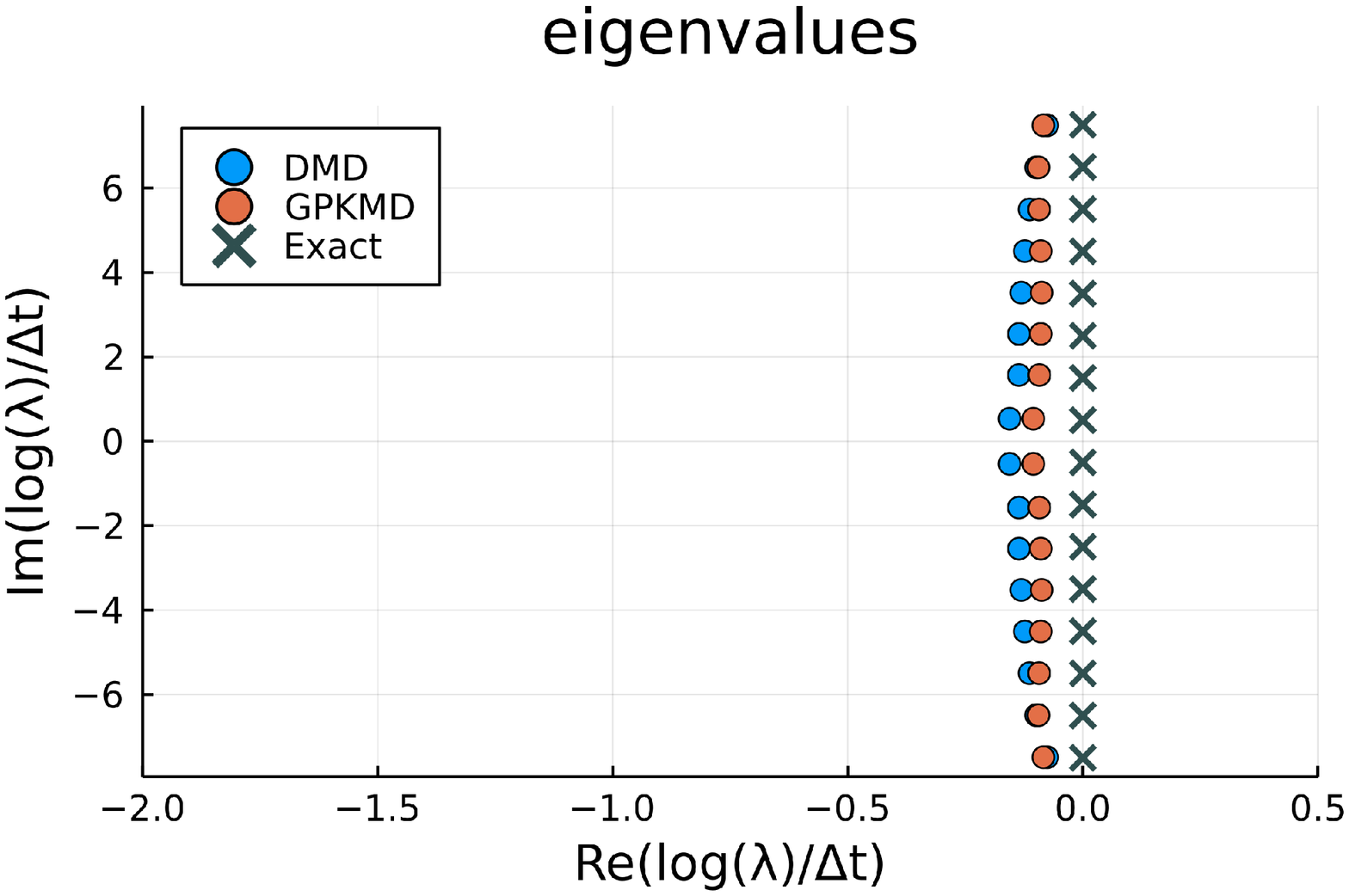}
        \subcaption{$\sigma = 0$}
        \label{subfig:sl0_eigvals}
    \end{minipage}
    \begin{minipage}{0.32\columnwidth}
        \centering
        \includegraphics[width = 1.0\columnwidth]{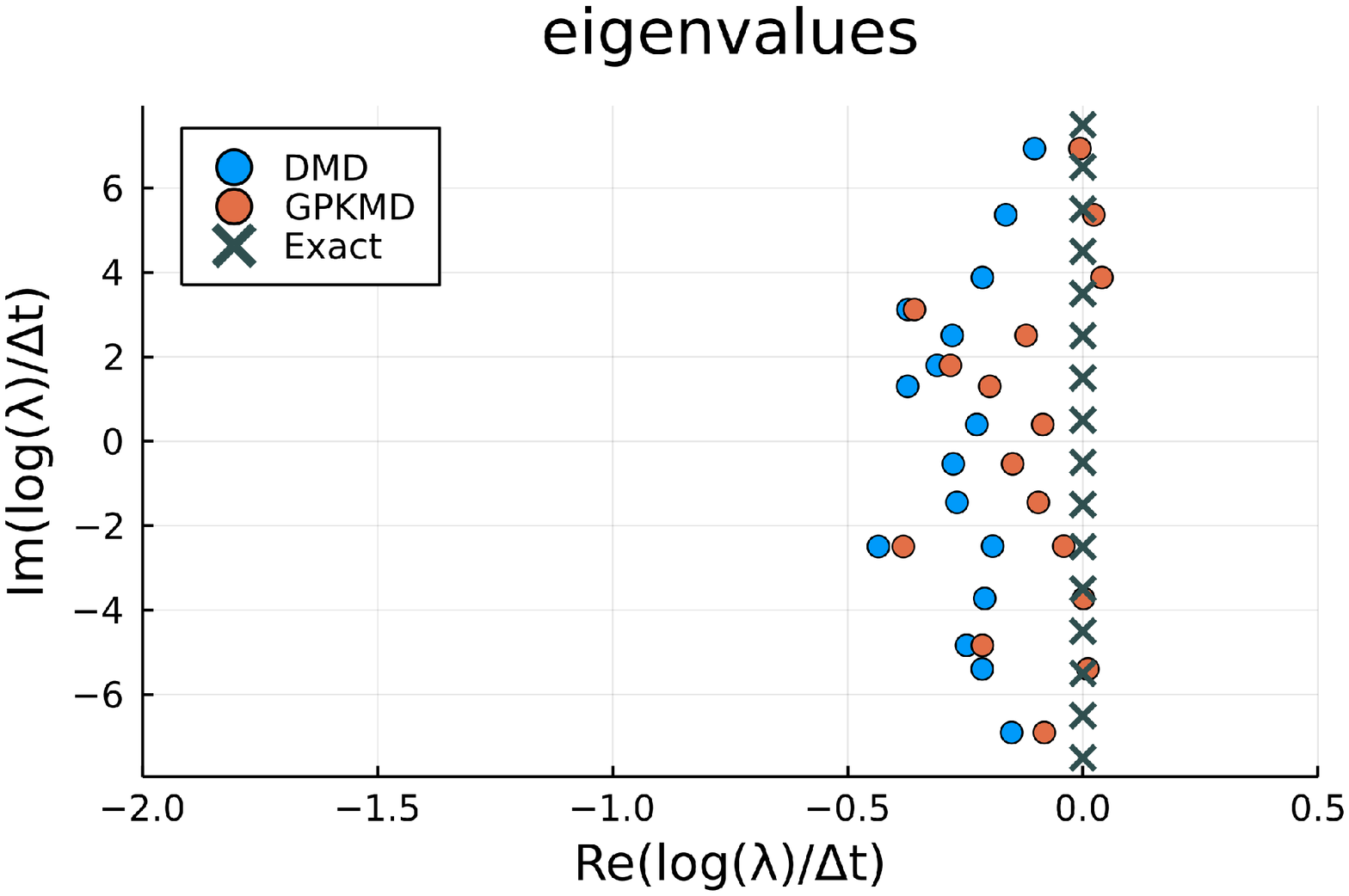}
        \subcaption{$\sigma = 0.01$}
        \label{subfig:sl001_eigvals}
    \end{minipage}
    \begin{minipage}{0.32\columnwidth}
        \centering
        \includegraphics[width = 1.0\columnwidth]{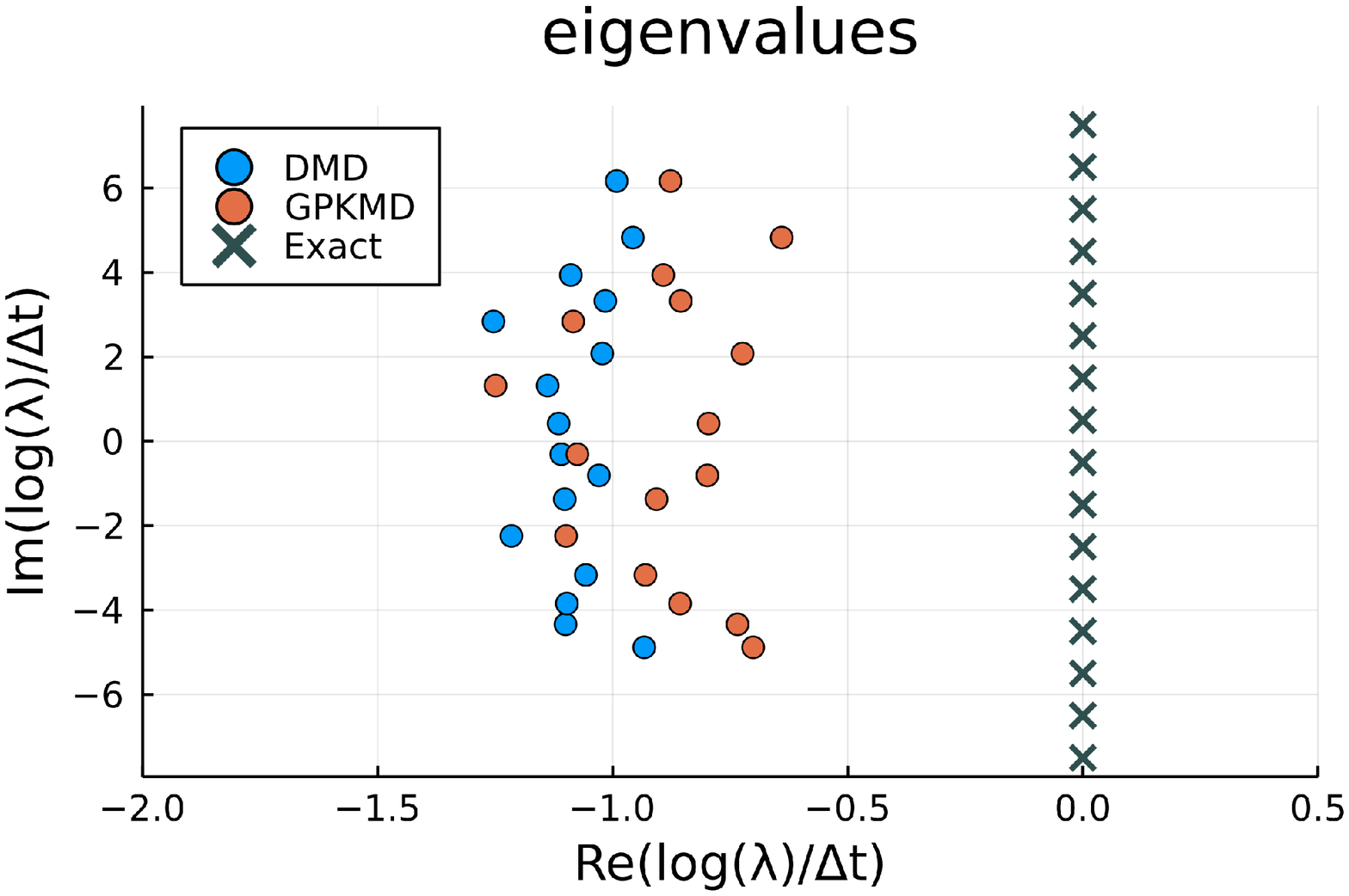}
        \subcaption{$\sigma = 0.2$}
        \label{subfig:sl02_eigvals}
    \end{minipage}
    \caption{
        Eigenvalues $\{\lambda^{\mathrm{cont}}_k\}$ estimated by DMD and GPKMD.
    }
    \label{fig:sl_eigvals}
\end{figure}

We generated data with $\delta = 0.5, \beta = \gamma = 1, \Delta t = 0.05$, and the data length $T = 751$.
As the observed data $\bs{Y} = (y_{dt})$ obtained through an observable, we employed
\begin{align}
    y_{dt} &= g_d(r_t, \theta_t) + \epsilon_{dt} = \exp(i d' \theta_t) + \epsilon_{dt}\\
    d' &= \left \{
    \begin{aligned}
        d' &= d - \lceil D/2 \rceil & \quad(d \mbox{~is odd})\\
        d' &= d/2 & \quad(d \mbox{~ is even})
    \end{aligned}
    \right.\\
    \epsilon_{dt} &\sim \CN(0, \sigma_y^2),
\end{align}
with input dimension $D = 35$ and noise levels $\sigma_y = 0, 0.01, 0.2.$
We used $K = 16$ modes, $P = 2$ latent dimensions, and $S = 50$ as the rank of the Gram matrices.
Figure \ref{fig:sl_latents} shows the latent variables estimated by PCA and GPKMD.
Although PCA and GPKMD estimates nearly the same trajectories for $\sigma_y = 0, 0.01$,
at the higher noise level $\sigma_y = 0.2,$ the latent variables of PCA are buried in the noise around the origin $\bs{x}_t = (0, 0)^\top$.
In contrast, GPKMD captures a contiguous and periodic trajectory around the origin for $\sigma_y = 0.2.$
The estimated Koopman eigenvalues corresponding to the continuous system
$\lambda^{\mathrm{cont}}_k = \log(\lambda_k) / \Delta t$ are shown in Figure \ref{fig:sl_eigvals}.
Note that the exact eigenvalues of the continuous system are known:
\begin{align}
    \lambda^{\mathrm{exact}}_{ln} &= -2l\delta + in\omega_0,\\
    \omega_0 &= \gamma - \beta \delta,
\end{align}
where $l \in \mathbb{L}$ and $n \in \mathbb{N}$ \citep{crnjaric-zic2020}.
As seen in Figure \ref{fig:sl_eigvals} and Table \ref{tab:sl_eigvals_error}
\footnote{Because $\mathrm{Im}(\lambda^{\mathrm{cont}}_k)$ is equal for DMD and GPKMD in our setting,
as discussed in Section \ref{discussion}, we only consider the real parts to obtain the errors.},
GPKMD estimates the Koopman eigenvalues more accurately than DMD.
The estimates of DMD tend to be biased as the noise level increases.
Meanwhile, though depending on initial values and hyperparameters, GPKMD is more robust than DMD for this dynamical system.

\begin{table}[t]
    \centering
    \caption{
        Absolute errors of the estimated eigenvalues
        $\lVert\mathrm{Re}(\bs{\lambda}^{\mathrm{exact}} - \bs{\lambda}^{\mathrm{cont}})\rVert$.
    }
    \label{tab:sl_eigvals_error}
    \begin{tabular}{c|ccc}
        \toprule
         ~     & $\sigma = 0$ & $\sigma = 0.01$ & $\sigma = 0.2$ \\
         \midrule
         DMD   & 0.49 & 1.06 & 4.32 \\
         GPKMD & 0.37 & 0.71 & 3.61 \\
         \bottomrule
    \end{tabular}
\end{table}

\subsection{Google Flu Trends}
\begin{figure*}[t]
    \centering
    \begin{minipage}{0.32\columnwidth}
        \centering
        \includegraphics[width = 1.0\columnwidth]{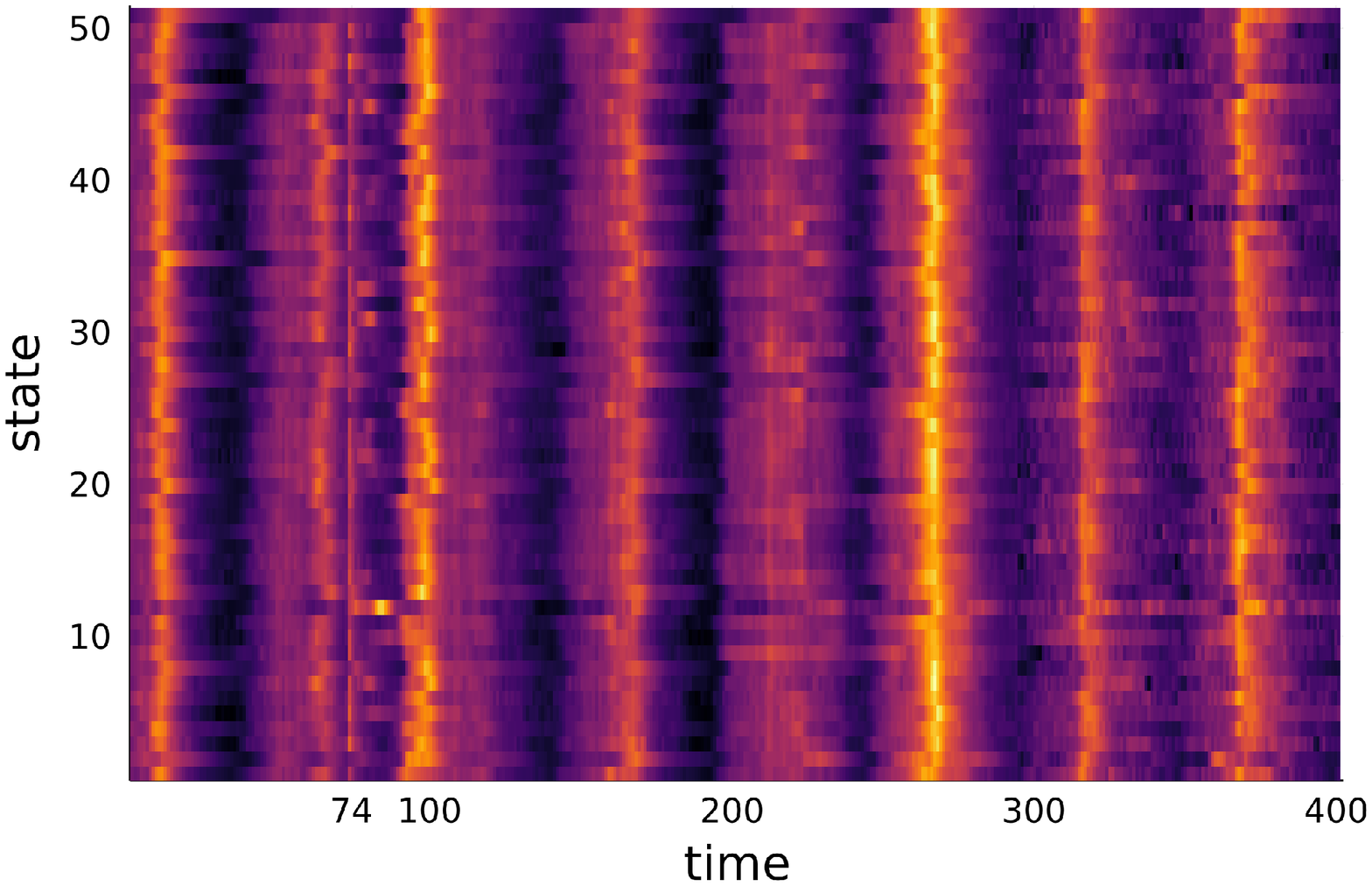}
        \subcaption{Input}
        \label{subfig:flu_obs}
    \end{minipage}
    \begin{minipage}{0.32\columnwidth}
        \centering
        \includegraphics[width = 1.0\columnwidth]{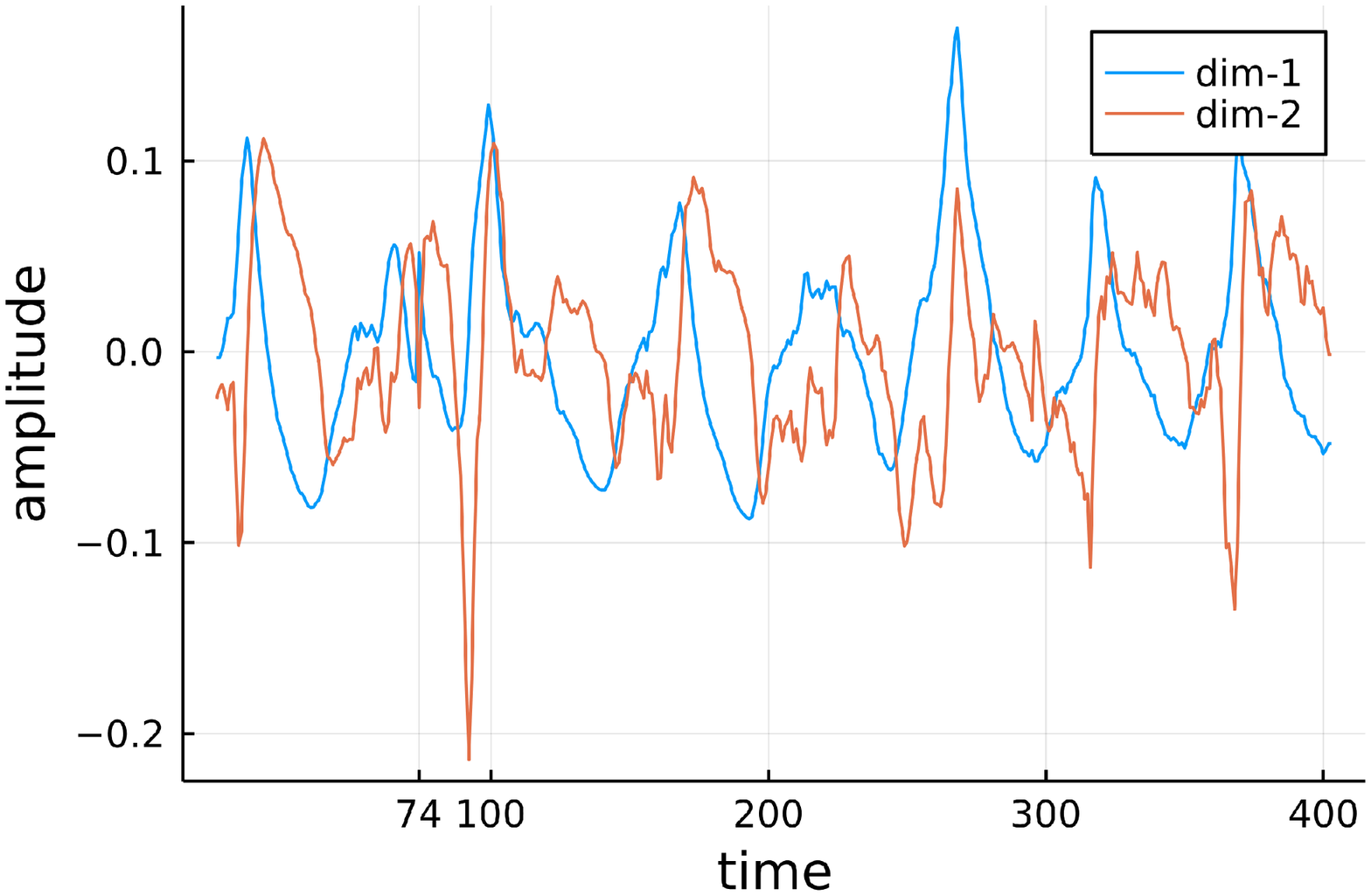}
        \subcaption{Latents (PCA)}
        \label{subfig:flu_latents_pca}
    \end{minipage}
    \begin{minipage}{0.32\columnwidth}
        \centering
        \includegraphics[width = 1.0\columnwidth]{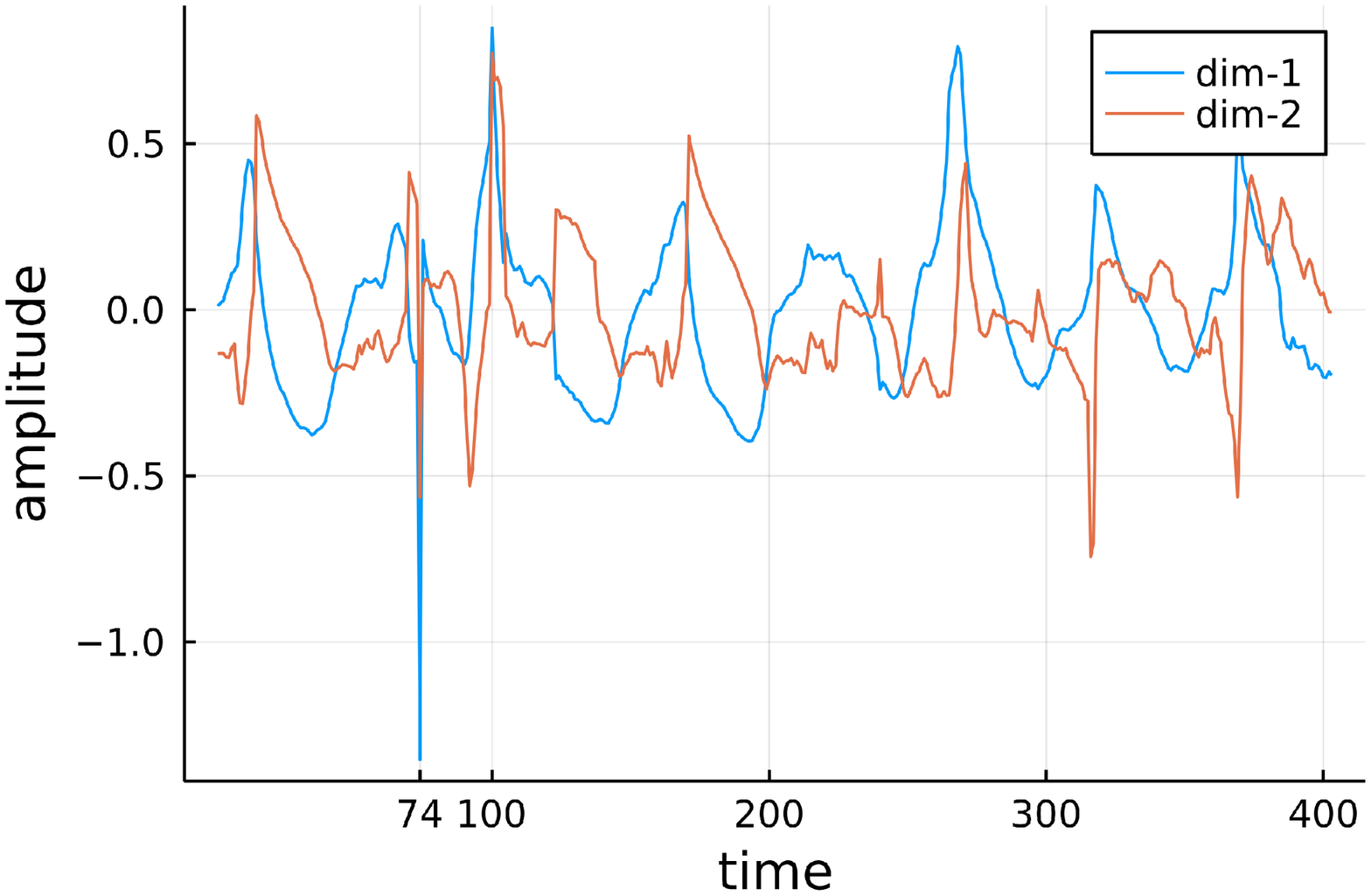}
        \subcaption{Latents (GPKMD)}
        \label{subfig:flu_latents_gpkmd}
    \end{minipage}
    \caption{
    \subref{subfig:flu_obs} Input from Google Flu Trends in the US.
    \subref{subfig:flu_latents_pca} Latent variables estimated by PCA.
    \subref{subfig:flu_latents_gpkmd} Latent variables estimated by GPKMD.
    }
    \label{fig:flu_latents}
\end{figure*}

Google has attempted to predict weekly spatiotemporal flu activity from query data of its search engine.
The project Google Flu Trends has been discontinued, but the predicted results are available
\footnote{The estimates can be accessed at \url{https://www.google.com/publicdata/explore?ds=z3bsqef7ki44ac_&hl=en&dl=en},
and the raw data is archived at \url{http://web.archive.org/web/*/http://www.google.org/flutrends/}.}.
\citet{proctor2015} analyzed the Google Flu Trends data by DMD, and we take a similar approach here.
We focused on the values in the US and extracted the interval from 2007--12--02 to 2015--08--09 to avoid missing data,
so that the input size was $D = 51, T = 402.$
Considering the nature of the data, we applied log-transformation before statewise standardization as preprocessing.
In the preprocessed input shown in Figure~\ref{subfig:flu_obs},
a rough periodicity can be observed. 
We set $K = 6$ modes, $P = 2$ latent dimensions, and $S = 50$ as the rank of the Gram matrices.\par

\begin{figure*}[t]
    \centering
    \begin{minipage}{1.0\columnwidth}
        \centering
        \includegraphics[width = 1.0\columnwidth]{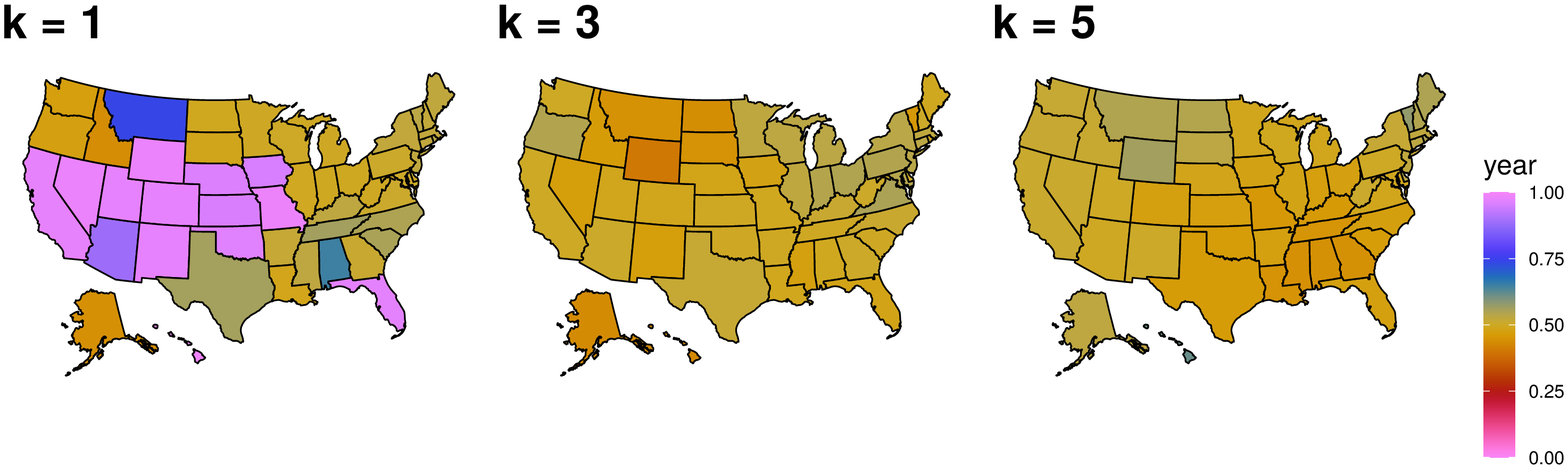}
        \subcaption{Phases of modes (DMD)}
        \label{subfig:flu_phase_dmd}
    \end{minipage}
    \\
    \begin{minipage}{1.0\columnwidth}
        \centering
        \includegraphics[width = 1.0\columnwidth]{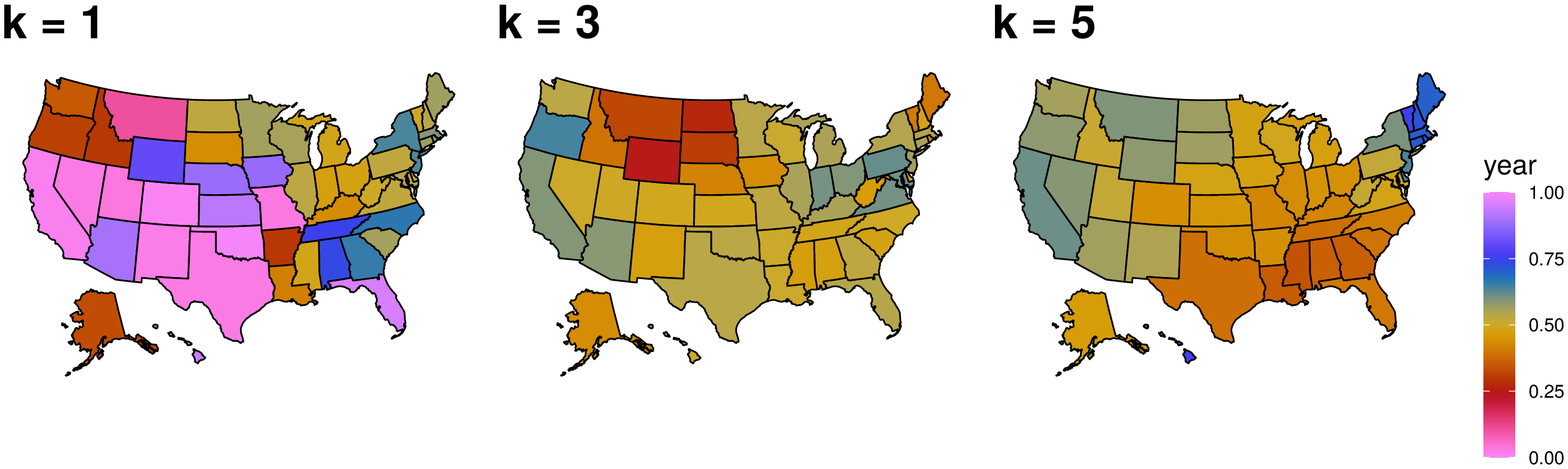}
        \subcaption{Phases of modes (GPKMD)}
        \label{subfig:flu_phase_gpkmd}
    \end{minipage}
    \caption{
    Phases of $1, 3,$ and $5$-th modes estimated by
    \subref{subfig:flu_phase_dmd} DMD and
    \subref{subfig:flu_phase_gpkmd} GPKMD.
    Each phase indicates the time of a year between 0 and 1.
    The corresponding frequencies are estimated as $0.068, 0.600, 1.426 ~\mbox{[1/year]}$ for $k = 1, 3, 5$, respectively.
    }
    \label{fig:flu_phases}
\end{figure*}

Figures \ref{subfig:flu_latents_pca} and \ref{subfig:flu_latents_gpkmd}
show the latent variables estimated by PCA and GPKMD, respectively.
Note that the latent variables estimated by PCA are used as the initial values of those estimated by GPKMD.
The latent variables estimated by GPKMD clearly show anomalous behavior at $t = 74,$ unlike those estimated by PCA.
An anomalous spike at $t = 74$ can also be observed in the original input (Figure \ref{subfig:flu_obs}),
but it does not appear to be outlying in the sense of i.i.d. observation.
$t = 74$ indicates the period between 2009--04--26 to 2009--05--02.
At the time, interestingly, the US was in turmoil due to the pandemic by new influenza A(H1N1).
In fact, WHO has raised the level of influenza pandemic alert to phase 4 on 2009--04--27,
and again raised to phase 5 on 2009--04--29 \citep{worldhealthorganization2013}.
The spike may reflect this social situation.
It is considered that the temporal structure and nonlinearity of GPKMD increase the sensitivity
to such temporally anomalous behavior.
In addition, the estimated modes $\{\bs{w}_k\}$ provide information about the phase shifts,
that is, the phase of the $k$-th mode in the $d$-th state is computed from $\mathrm{arg}~ w_{dk}$.
Suppose that $\mathrm{arg}~ w_{dk}$ is wrapped to $[0, 2\pi)$,
then $\mathrm{arg}~ w_{dk} ~/~ 2\pi \in [0, 1)$ expresses the shift within a year.
Figure \ref{fig:flu_phases} shows the phases of the modes corresponding to the indices $k = 1, 3, 5,$ estimated by DMD and GPKMD.
The modes indexed by even numbers are omitted because they have conjugate elements of odd numbers.
The first modes of DMD and GPKMD indicate some state clusters.
We also find a clustered relationship in the northern states at $k = 3$ and
a gradual slope from southeast to northwest at $k = 5$.
Such phase structures are considered to reflect seasonal transitions of flu trends.
Notably a similar smooth phase transition of a dynamic mode has also been reported in a previous work \citep{proctor2015},
but the transition is more pronounced for our method.

\section{Discussion and Conclusion}
\label{discussion}
In this study, we developed a nonlinear probabilistic generative model of KMD
based on unsupervised GP,
and we also proposed its efficient inference scheme via low-rank approximations of covariance matrices.
Our method, named GPKMD, is advantageous in terms of the comprehensiveness of the parameter set to be estimated.
Since each quantity in KMD \eqref{eq:kmd} is physically meaningful,
the comprehensiveness of GPKMD directly means that rich information can be obtained.
We also examined the scalability of GPKMD in Section \ref{scale}.
By exploiting the properties of the Kronecker product and low-rank approximations of matrices,
we markedly reduced the computational complexity from $\mathcal{O}(D^3 + T^3)$ to $\mathcal{O}(DK^2 + TS^2)$,
where $K \ll D$ and $S \ll T$.
In Section \ref{ex}, we applied GPKMD to synthetic and real datasets and interpreted the results
from different aspects through the estimated
Koopman eigenvalues $\{\lambda_k\}$, Koopman modes $\{\bs{w}_k\}$, and latent variables $\{\bs{x}_t\}$.\par
This study has some limitations, and some future works are suggested.
In this work, we did not show the estimated eigenfunctions $\{\phi_k\}$.
The eigenfunctions are implicitly determined with the kernel function and the latent variables in our model,
but their explicit estimates are intractable.
Approximated GPs with finite-dimensional features such as \textit{random Fourier features} \citep{rahimi2007} are possible approaches to obtain explicit expressions of the esimated eigenfunctions.
Although we employed gradient-based MAP estimation in Section \ref{ex},
a credible interval estimation of GPKMD will lead to an uncertainty-incorporated interpretation of results.
The \textit{sparse variational Gaussian process} (SVGP) is a well-known variational Bayesian method for learning GPs,
which maximizes the ELBO instead of the marginalized posterior~\citep{titsias2009, titsias2010}.
While connections between the Nystr{\"o}m method and SVGP is studied recently
\citep{wild2021},
a variational inference method for GPKMD still should be developed as future works.
There is also difficulty in learning the Koopman eigenvalues $\{\lambda_k\}$.
The angle of the $k$-th eigenvalue $\mathrm{arg}~ \lambda_k$ corresponds to the frequency of the $k$-th mode.
In \eqref{eq:lik}, however, the eigenvalues are included in the form $\bs{\Lambda} \bs{\Lambda}^\ast = \diag(|\lambda_k|^2)$;
hence, the angles $\{\mathrm{arg}~ \lambda_k\}$ do not affect the likelihood.
In addition, the gradient of the likelihood \eqref{eq:lik} w.r.t. $\lambda_k$ is proportional to $\lambda_k$ itself,
and the angle does not vary from the initial value during gradient-based learning.
In the examples in Section \ref{ex}, we practically use the DMD estimates of $\{\lambda_k\}$
to alleviate this difficulty, but how to determine the angles remains a problem.

\subsection*{Acknowledgement}
We thank an anonymous reviewer for insightful comments and suggestions.
Part of this work is supported by JST CREST JPMJCR1761, JPMJCR2015, JST JPMJFS2136, JSPS KAKENHI 19K12111.

\bibliographystyle{apalike}
\bibliography{NECO-D-22-00065-Bib}

\appendix

\section{Appendix}
\label{appendix}

\subsection{Properties of Kronecker Product and Vec Operator}
We introduce some properties of the Kronecker product and $\vect$ operator  for simplicity in the calculations below:
\begin{align}
    \label{eq:kron1}
    (\bs{A} \otimes \bs{B}) (\bs{C} \otimes \bs{D}) &= (\bs{A}\bs{C}) \otimes (\bs{B}\bs{D}),\\
    \label{eq:kron2}
    \vect(\bs{A}\bs{B}\bs{C}) &= (\bs{C}^\top \otimes \bs{A}) \vect(\bs{B}),\\
    \label{eq:kron3}
    \tr(\bs{A} \otimes \bs{B}) &= \tr(\bs{A})\tr(\bs{B}),\\
    \vect(\bs{A})^\ast \vect(\bs{B}) &= \tr(\bs{A}^\ast \bs{B}).
\end{align}

\subsection{Deriving Marginal Likelihood of GPKMD}
From each $\CN(\cdot, \cdot)$ in \eqref{eq:pre_lik} and the prior $p(b_{kl}) = \CN(0, \sigma^2_b)$,
we can marginalize out the coefficients $\{b_{kl}\}$ analytically.
Considering the joint marginal likelihood for the first $\CN(\cdot, \cdot)$ in \eqref{eq:pre_lik},
we have
\begin{align}
    p(\bs{Y} | \bs{X}, \bs{W}, \sigma^2) 
    &= \int \prod^T_{t=1} p(\bs{y}_{t} | \{\bs{x}_t\}, \{\lambda_k\}, \{\bs{w}_k\}, \{b_{kl}\}, \sigma^2) \cdot \prod_{k, l} p(b_{kl}) db_{kl}\\
    \label{eq:app1}
    &= \int \CN(\vect(\bs{Y}) | \vect(\bs{W}\bs{B}\bs{\Psi}_1), \sigma^2 \bs{I}) \CN(\vect(\bs{B}) | \bs{0}, \sigma^2_b \bs{I}) d\bs{B},\quad
\end{align}
where $\bs{\Psi}_1 = (\bs{\psi}(\bs{x}_1), \bs{\psi}(\bs{x}_2), \ldots)$ and $\bs{B}$ is the matrix whose $(k, l)$-th element is $b_{kl}$.
Using the relations \eqref{eq:kron1} and \eqref{eq:kron2}, we find that the integrand in \eqref{eq:app1} is proportional to 
\begin{align}
    &\CN(\vect(\bs{Y}) | \vect(\bs{W}\bs{B}\bs{\Psi}_1), \sigma^2 \bs{I}) \CN(\vect(\bs{B}) | \bs{0}, \sigma^2_b \bs{I})\\
    &\quad\propto \exp \left (-\sigma^{-2} \lVert \vect(\bs{Y}) - \vect(\bs{W}\bs{B}\bs{\Psi}_1)) \rVert^2 - \sigma^2_b \lVert \vect(\bs{B}) \rVert^2 \right )\\
    \label{eq:app2}
    &\quad= \exp \left \{ -\sigma^{-2} \left ( \lVert \vect(\bs{Y}) \rVert^2 + \lVert \vect(\bs{B}) + \vect(\bar{\bs{B}}) \rVert^2_{\bs{\Sigma}^{-1}_B} - \lVert \vect(\bar{\bs{B}})\rVert^2_{\bs{\Sigma}^{-1}_B} \right ) \right \}, \quad
\end{align}
where
\begin{align}
    \bs{\Sigma}^{-1}_B &= (\bs{\Psi}_1 \bs{\Psi}^\top_1) \otimes (\bs{W}^\ast \bs{W}) + \sigma^2 \sigma^{-2}_b \bs{I},\\
    \vect(\bar{\bs{B}}) &= \bs{\Sigma}_B \vect(\bs{W}^\ast \bs{Y} \bs{\Psi}^\top_1),\\
    \lVert \bs{z} \lVert^2_{\bs{\Sigma}^{-1}_B} &= \bs{z}^\ast \bs{\Sigma}^{-1}_B \bs{z}.
\end{align}
Since \eqref{eq:app2} is the squared exponential form w.r.t. $\vect(\bs{B})$,
the integral \eqref{eq:app1} can be evaluated as
\begin{align}
    &\int \CN(\vect(\bs{Y}) | \vect(\bs{W}\bs{B}\bs{\Psi}_1), \sigma^2 \bs{I}) \CN(\vect(\bs{B}) | \bs{0}, \sigma^2_b \bs{I}) d\bs{B} \\
    \label{eq:app3}
    &\quad\propto \exp \left \{ -\sigma^{-2} \left ( \lVert \vect(\bs{Y}) \rVert^2 - \lVert \vect(\bar{\bs{B}})\rVert^2_{\bs{\Sigma}^{-1}_B} \right ) \right \},
\end{align}
and this should also be Gaussian w.r.t. $\vect(\bs{Y})$.
Here, applying the Woodbury identity and \eqref{eq:kron2}, we obtain
\begin{align}
    \bs{\Sigma}_B &= ((\bs{\Psi}_1 \bs{\Psi}^\top_1) \otimes (\bs{W}^\ast \bs{W}) + \sigma^2 \sigma^{-2}_b \bs{I})^{-1}\\
    &= \sigma^{-2}\sigma^2_b \{\bs{I} - \sigma^2_b (\bs{\Psi}_1 \otimes \bs{W}^\ast) \bs{\Sigma}^{-1}_Y (\bs{\Psi}^\top_1 \otimes \bs{W}) \},
\end{align}
where we define
\begin{align}
    \bs{\Sigma}_Y &= \sigma^2\bs{I} + \sigma^2_b \bs{K}_1 \otimes (\bs{W} \bs{W}^\ast),\\
    \bs{K}_1 &= \bs{\Psi}^\top_1 \bs{\Psi}_1.
\end{align}
Then, $\lVert \vect(\bar{\bs{B}})\rVert^2_{\bs{\Sigma}^{-1}_B}$ in \eqref{eq:app3} can be simplified to
\begin{align}
    &\lVert \vect(\bar{\bs{B}})\rVert^2_{\bs{\Sigma}^{-1}_B}\\
    &\quad= \vect(\bs{W}^\ast \bs{Y} \bs{\Psi}^\top_1)^\ast \bs{\Sigma}_B \vect(\bs{W}^\ast \bs{Y} \bs{\Psi}^\top_1) \\
    &\quad= \sigma^{-2}\sigma^2_b \lVert \vect(\bs{W}^\ast \bs{Y} \bs{\Psi}^\top_1) \rVert^2
    -\sigma^{-2}\sigma^2_b \lVert (\bs{\Psi}^\top_1 \otimes \bs{W}) \vect(\bs{W}^\ast \bs{Y} \bs{\Psi}^\top_1) \rVert^2_{\sigma^{2}_b \bs{\Sigma}^{-1}_Y} \\
    &\quad= \sigma^{-2}\sigma^2_b\tr(\bs{\Psi}_1 \bs{Y}^\ast \bs{W}\bs{W}^\ast \bs{Y} \bs{\Psi}^\top_1)
    -\sigma^{-2}\sigma^2_b \lVert \vect(\bs{W}\bs{W}^\ast \bs{Y} \bs{K}_1) \rVert^2_{\sigma^{2}_b \bs{\Sigma}^{-1}_Y}\\
    &\quad= \sigma^{-2}\sigma^2_b \vect(\bs{W}\bs{W}^\ast \bs{Y})^\ast \vect(\bs{Y} \bs{K}_1)
    -\sigma^{-2}\sigma^2_b \lVert (\bs{K}_1 \otimes (\bs{W}\bs{W}^\ast) ) \vect(\bs{Y}) \rVert^2_{\sigma^{2}_b \bs{\Sigma}^{-1}_Y}\\
    &\quad= \sigma^{-2}\sigma^2_b \vect(\bs{Y})^\ast \\
    &\qquad\times [\bs{K}_1 \otimes (\bs{W}\bs{W}^\ast ) - \sigma^2_b(\bs{K}_1 \otimes (\bs{W}\bs{W}^\ast) )\bs{\Sigma}^{-1}_Y(\bs{K}_1 \otimes (\bs{W}\bs{W}^\ast) ) ] \vect(\bs{Y})\\
    &\quad= \vect(\bs{Y})^\ast (\bs{I} - \sigma^{2}\bs{\Sigma}^{-1}_Y) \vect(\bs{Y}),
\end{align}
where we use the exact relation $\bs{A} - \bs{A}(\bs{A} + \bs{B})^{-1} \bs{A} = \bs{B} - \bs{B}(\bs{A} + \bs{B})^{-1} \bs{B}$ for the rightmost transform.
Now, the (unnormalized) marginal likelihood \eqref{eq:app3} becomes
\begin{align}
    p(\bs{Y} | \bs{X}, \bs{W}, \sigma^2)
    &\propto \exp \left \{ -\sigma^{-2} \left ( \lVert \vect(\bs{Y}) \rVert^2 - \lVert \vect(\bar{\bs{B}})\rVert^2_{\bs{\Sigma}^{-1}_B} \right ) \right \}\\
    &= \exp \left \{ -\sigma^{-2} \left ( \vect(\bs{Y})^\ast \vect(\bs{Y}) - \vect(\bs{Y})^\ast (\bs{I} - \sigma^{2}\bs{\Sigma}^{-1}_Y) \vect(\bs{Y}) \right ) \right \}\\
    &= \exp \left (-\vect(\bs{Y})^\ast \bs{\Sigma}^{-1}_Y \vect(\bs{Y}) \right ),
\end{align}
so that
$p(\bs{Y} | \bs{X}, \bs{W}, \sigma^2, \sigma^2_b) = \CN(\vect(\bs{Y}) | \bs{0}, \bs{\Sigma}_Y)$.
Applying a similar manner to the second $\CN(\cdot, \cdot)$ in \eqref{eq:pre_lik}, we can finally obtain the marginal likelihood of GPKMD \eqref{eq:lik}.

\subsection{Derivatives of the Marginal Likelihood}
In Section \ref{scale}, we show that the low-rank approximations for the covariance matrices reduce the computational cost of evaluating the GPKMD likelihood.
Similarly, we can evaluate derivatives of the likelihood efficiently.
For the complex-valued parameters of GPKMD,
we define the complex gradient of $f: \mathbb{C}^D \to \mathbb{R}$ w.r.t. $\bs{\theta} \in \mathbb{C}^D$ as
\begin{align}
    \nabla_{\bs{\theta}} f(\bs{\theta}) = \cfrac{\partial f(\bs{\theta})}{\partial \re(\bs{\theta})} + i\cfrac{\partial f(\bs{\theta})}{\partial \im(\bs{\theta})} .
\end{align}
In general, we consider the cost function
\begin{align}
    \ell(\bs{\theta}^g, \bs{\theta}^h)
    &= \log \det(\sigma^2 \bs{I} + \bs{G}(\bs{\theta}^g) \otimes \bs{H}(\bs{\theta}^h))\\
    \label{eq:app4}
    &\quad - \vect(\bs{Y})^\ast (\sigma^2 \bs{I} + \bs{G}(\bs{\theta}^g) \otimes \bs{H}(\bs{\theta}^h))^{-1} \vect(\bs{Y}),
\end{align}
where $\bs{G}$ and $\bs{H}$ are positive semidefinite and $\bs{\theta}^g$ and $\bs{\theta}^h$ are the parameter vectors to be learned.
As introduced in Section \ref{scale}, suppose we obtain low-rank representations such that
$\bs{G} \approx \bs{U}_G \bs{\Sigma}^2_G \bs{U}^\ast_G$ and $\bs{H} \approx \bs{U}_H \bs{\Sigma}^2_H \bs{U}^\ast_H$ by SVD.
Then, the Woodbury identity enables the following approximation:
\begin{align}
    &(\sigma^2 \bs{I} + \bs{G} \otimes \bs{H})^{-1} \\
    \label{eq:app5}
    &\quad \approx \sigma^{-2} \{\bs{I} - [(\bs{U}_{G} \bs{\Sigma}_{G}) \otimes (\bs{U}_H \bs{\Sigma}_H)]
    (\sigma^2 \bs{I} + \bs{\Sigma}^2_{G} \otimes \bs{\Sigma}^2_H)^{-1}
    [(\bs{U}_{G} \bs{\Sigma}_{G}) \otimes (\bs{U}_H \bs{\Sigma}_H)]^\ast\}.
\end{align}

\subsubsection*{Derivatives w.r.t. $\bs{\theta}^g$}
The derivative of the cost function $\ell(\bs{\theta}^g, \bs{\theta}^h)$ w.r.t. $\theta^g_i$ is
\begin{align}
    \nabla_{\theta^g_i}\ell
    &= -\tr\{(\sigma^2 \bs{I} + \bs{G} \otimes \bs{H})^{-1} (\nabla_{\theta^g_i}\bs{G} \otimes \bs{H})\}\\
    \label{eq:app6}
    &\quad + \vect(\bs{Y})^\ast (\sigma^2 \bs{I} + \bs{G} \otimes \bs{H})^{-1} (\nabla_{\theta^g_i}\bs{G} \otimes \bs{H})
    (\sigma^2 \bs{I} + \bs{G} \otimes \bs{H})^{-1}\vect(\bs{Y}).\quad
\end{align}
The first term in \eqref{eq:app6} can be approximated by
\begin{align}
    &\tr\{(\sigma^2 \bs{I} + \bs{G} \otimes \bs{H})^{-1} (\nabla_{\theta^g_i}\bs{G} \otimes \bs{H})\}\\
    &\quad\approx \sigma^{-2} \tr(\nabla_{\theta^g_i}\bs{G} \otimes \bs{H})
    - \sigma^{-2} \tr\{[(\bs{U}_{G} \bs{\Sigma}_{G}) \otimes (\bs{U}_H \bs{\Sigma}_H)]\\
    &\qquad\times (\sigma^2 \bs{I} + \bs{\Sigma}^2_{G} \otimes \bs{\Sigma}^2_H)^{-1}
    [(\bs{U}_{G} \bs{\Sigma}_{G}) \otimes (\bs{U}_H \bs{\Sigma}_H)]^\ast (\nabla_{\theta^g_i}\bs{G} \otimes \bs{H})\}\\
    &\quad= \sigma^{-2} \tr(\nabla_{\theta^g_i}\bs{G})\tr(\bs{H})
    - \sigma^{-2} \tr\{(\sigma^2 \bs{I} + \bs{\Sigma}^2_{G} \otimes \bs{\Sigma}^2_H)^{-1}\\
    &\qquad\times [(\bs{\Sigma}_{G} \bs{U}^\ast_{G} \nabla_{\theta^g_i}\bs{G} \bs{U}_{G} \bs{\Sigma}_{G}) \otimes \bs{\Sigma}^4_H]\}\\
    &\quad= \sigma^{-2} \tr(\nabla_{\theta^g_i}\bs{G})\tr(\bs{\Sigma}^2_H)\\
    &\qquad- \sigma^{-2} \diag\{(\sigma^2 \bs{I} + \bs{\Sigma}^2_{G} \otimes \bs{\Sigma}^2_H)^{-1}\}^\top
    \diag\{(\bs{\Sigma}_{G} \bs{U}^\ast_{G} \nabla_{\theta^g_i}\bs{G} \bs{U}_{G} \bs{\Sigma}_{G}) \otimes \bs{\Sigma}^4_H\}\\
    &\quad= \sigma^{-2} \tr(\nabla_{\theta^g_i}\bs{G})\tr(\bs{\Sigma}^2_H)\\
    &\qquad- \sigma^{-2} \diag\{(\sigma^2 \bs{I} + \bs{\Sigma}^2_{G} \otimes \bs{\Sigma}^2_H)^{-1}\}^\top
    \{\diag(\bs{\Sigma}_{G} \bs{U}^\ast_{G} \nabla_{\theta^g_i}\bs{G} \bs{U}_{G} \bs{\Sigma}_{G}) \otimes \diag(\bs{\Sigma}^4_H)\},
\end{align}
where we use \eqref{eq:kron3} and $\tr(\bs{D}\bs{A}) = \diag(\bs{D})^\top \diag(\bs{A})$ for any diagonal matrix $\bs{D}$.
Note that if $\bs{G}$ is a Gram matrix of latent variables $\bs{X} = (\bs{x}_1, \bs{x}_2, \ldots, \bs{x}_T)^\top$,
i.e., $\bs{G} = (k(\bs{x}_i, \bs{x}_j))_{ij}~ (= \bs{K}_1 \mbox{ in \eqref{eq:lik}})$,
then the elements of $\nabla_{x_{pi}}\bs{G}$ become zeros except for the
$i$-th row and column, and $\tr(\nabla_{x_{pi}}\bs{G}) = 0$.
In such a case, further simplification is possible:
\begin{align}
    &\tr\{(\sigma^2 \bs{I} + \bs{G} \otimes \bs{H})^{-1} (\nabla_{x_{pi}}\bs{G} \otimes \bs{H})\}\\
    &\quad\approx - 2\sigma^{-2} \diag\{(\sigma^2 \bs{I} + \bs{\Sigma}^2_{G} \otimes \bs{\Sigma}^2_H)^{-1}\}^\top \\
    &\qquad\times \{[(\bs{U}^\ast_{G} \nabla_{x_{pi}}\bs{G}_{:i}) \odot \bs{U}_{G, :i} \odot \diag(\bs{\Sigma}^2_{G})] \otimes \diag(\bs{\Sigma}^4_H)\},
\end{align}
where $\odot$ denotes the Hadamard product and
$\nabla_{x_{pi}}\bs{G}_{:i}$ and $\bs{U}_{G, :i}$ are the $i$-th column vectors of $\nabla_{x_{pi}}\bs{G}$ and $\bs{U}_G$, respectively.\par

We next consider the second term in \eqref{eq:app6}.
By defining the transformed data onto the lower dimension
\begin{align}
    \vect(\tilde{\bs{Y}}) = (\sigma^2 \bs{I} + \bs{\Sigma}^2_{G} \otimes \bs{\Sigma}^2_H)^{-1}
    \vect(\bs{\Sigma}_H \bs{U}^\ast_H \bs{Y} \bs{U}_G \bs{\Sigma}_G),
\end{align}
we obtain the following approximation:
\begin{align}
    &\vect(\bs{Y})^\ast (\sigma^2 \bs{I} + \bs{G} \otimes \bs{H})^{-1} (\nabla_{\theta^g_i}\bs{G} \otimes \bs{H})
        (\sigma^2 \bs{I} + \bs{G} \otimes \bs{H})^{-1}\vect(\bs{Y})\\
    &\quad\approx \sigma^{-4} \{\vect(\bs{Y}) - [(\bs{U}_{G} \bs{\Sigma}_{G}) \otimes (\bs{U}_H \bs{\Sigma}_H)]\vect(\bs{\tilde{Y}})\}^\ast
    [\nabla_{\theta^g_i}\bs{G} \otimes (\bs{U}_H \bs{\Sigma}^2_H \bs{U}^\ast_H)]\\
    &\qquad\times \{\vect(\bs{Y}) - [(\bs{U}_{G} \bs{\Sigma}_{G}) \otimes (\bs{U}_H \bs{\Sigma}_H)]\vect(\bs{\tilde{Y}})\}\\
    &\quad= \sigma^{-4}\vect(\bs{Y})^\ast [\nabla_{\theta^g_i}\bs{G} \otimes (\bs{U}_H \bs{\Sigma}^2_H \bs{U}^\ast_H)] \vect(\bs{Y})\\
    &\qquad - \sigma^{-4} \vect(\bs{\tilde{Y}})^\ast [(\bs{\Sigma}_G \bs{U}^\ast_G \nabla_{\theta^g_i}\bs{G}) \otimes (\bs{\Sigma}^3_H \bs{U}^\ast_H)] \vect(\bs{Y}) \\
    &\qquad - \sigma^{-4} \vect(\bs{Y})^\ast [(\nabla_{\theta^g_i}\bs{G} \bs{U}_G \bs{\Sigma}_G) \otimes (\bs{U}_H \bs{\Sigma}^3_H )] \vect(\bs{\tilde{Y}}) \\
    &\qquad + \sigma^{-4} \vect(\bs{\tilde{Y}})^\ast [(\bs{\Sigma}_G \bs{U}^\ast_G \nabla_{\theta^g_i}\bs{G} \bs{U}_G \bs{\Sigma}_G) \otimes \bs{\Sigma}^4_H)] \vect(\bs{\tilde{Y}})\\
    &\quad= \sigma^{-4}\tr(\bs{\Sigma}^2_H\bs{U}^\ast_H\bs{Y} \nabla_{\theta^g_i}\bs{G}^\top \bs{Y}^\ast \bs{U}_H)
    - \sigma^{-4} \tr(\bs{\Sigma}^3_H \bs{U}^\ast_H \bs{Y} \nabla_{\theta^g_i}\bs{G}^\top \overline{\bs{U}_G} \bs{\Sigma}_G \tilde{\bs{Y}}^\ast)\\
    &\qquad - \sigma^{-4} \tr(\tilde{\bs{Y}} \bs{\Sigma}_G \bs{U}^\top_G \nabla_{\theta^g_i}\bs{G}^\top \bs{Y}^\ast \bs{U}_H \bs{\Sigma}^3_H )
    + \sigma^{-4}\tr(\bs{\Sigma}^4_H \tilde{\bs{Y}} \bs{\Sigma}_G \bs{U}^\top_G \nabla_{\theta^g_i}\bs{G}^\top \overline{\bs{U}_G} \bs{\Sigma}_G \tilde{\bs{Y}}^\ast ),
\end{align}
where $\overline{\bs{U}_H}$ denotes the conjugate matrix without the transpose of $\bs{U}_H$.
Furthermore, in the particular case where $\bs{G}$ is a Gram matrix consisting of $\bs{X} = (\bs{x}_1, \bs{x}_2, \ldots, \bs{x}_T)^\top$,
we have
\begin{align}
    &\vect(\bs{Y})^\ast (\sigma^2 \bs{I} + \bs{G} \otimes \bs{H})^{-1} (\nabla_{x_{pi}}\bs{G} \otimes \bs{H})
        (\sigma^2 \bs{I} + \bs{G} \otimes \bs{H})^{-1}\vect(\bs{Y})\\
    &\quad\approx 2\sigma^{-4} \re \{ \nabla_{x_{pi}}\bs{G}^\top_{:i} \bs{Y}^\ast \bs{U}_H \bs{\Sigma}^2_H \bs{U}^\ast_H \bs{Y}_{:t}
    - \nabla_{x_{pi}}\bs{G}^\top_{:i} \bs{U}_G \bs{\Sigma}_G \tilde{\bs{Y}}^\ast \bs{\Sigma}^3_H \bs{U}^\ast_H \bs{Y}_{:t}\\
    &\qquad - \bs{U}^\top_{G, i:} \bs{\Sigma}_G \tilde{\bs{Y}}^\ast \bs{\Sigma}^3_H \bs{U}_H \bs{Y} \nabla_{x_{pi}}\bs{G}_{:i}
    + \nabla_{x_{pi}}\bs{G}^\top_{:i} \bs{U}_G \bs{\Sigma}_G \tilde{\bs{Y}}^\ast \bs{\Sigma}^4_H \tilde{\bs{Y}} \bs{\Sigma}_G \bs{U}_{G, i:}\}.sl_eigvals_sigma0
\end{align}

\subsubsection*{Derivatives w.r.t. $\bs{\theta}^h$}
The derivative of the cost function $\ell(\bs{\theta}^g, \bs{\theta}^h)$ w.r.t. $\theta^h_i$ is
\begin{align}
    \nabla_{\theta^h_i}\ell
    &= -\tr\{(\sigma^2 \bs{I} + \bs{G} \otimes \bs{H})^{-1} (\bs{G} \otimes \nabla_{\theta^h_i}\bs{H})\}\\
    \label{eq:app7}
    &\quad + \vect(\bs{Y})^\ast (\sigma^2 \bs{I} + \bs{G} \otimes \bs{H})^{-1} (\bs{G} \otimes \nabla_{\theta^h_i}\bs{H})
    (\sigma^2 \bs{I} + \bs{G} \otimes \bs{H})^{-1}\vect(\bs{Y}).\quad
\end{align}
We can approximate the first term in \eqref{eq:app7} as
\begin{align}
    &\tr\{(\sigma^2 \bs{I} + \bs{G} \otimes \bs{H})^{-1} (\bs{G} \otimes \nabla_{\theta^h_i}\bs{H})\}\\
    &\quad\approx \sigma^{-2} \tr(\bs{G} \otimes \nabla_{\theta^h_i}\bs{H})
    - \sigma^{-2} \tr\{[(\bs{U}_{G} \bs{\Sigma}_{G}) \otimes (\bs{U}_H \bs{\Sigma}_H)]\\
    &\qquad\times (\sigma^2 \bs{I} + \bs{\Sigma}^2_{G} \otimes \bs{\Sigma}^2_H)^{-1}
    [(\bs{U}_{G} \bs{\Sigma}_{G}) \otimes (\bs{U}_H \bs{\Sigma}_H)]^\ast (\bs{G} \otimes \nabla_{\theta^h_i}\bs{H})\}\\
    &\quad= \sigma^{-2} \tr(\bs{G})\tr(\nabla_{\theta^h_i}\bs{H})
    - \sigma^{-2} \tr\{(\sigma^2 \bs{I} + \bs{\Sigma}^2_{G} \otimes \bs{\Sigma}^2_H)^{-1}\\
    &\qquad\times [\bs{\Sigma}^4_G \otimes (\bs{\Sigma}_{H} \bs{U}^\ast_{H} \nabla_{\theta^h_i}\bs{H} \bs{U}_{H} \bs{\Sigma}_{H})]\}\\
    &\quad= \sigma^{-2} \tr(\bs{\Sigma}^2_G)\tr(\nabla_{\theta^h_i}\bs{H})\\
    &\qquad- \sigma^{-2} \diag\{(\sigma^2 \bs{I} + \bs{\Sigma}^2_{G} \otimes \bs{\Sigma}^2_H)^{-1}\}^\top
    \diag\{\bs{\Sigma}^4_G \otimes (\bs{\Sigma}_{H} \bs{U}^\ast_{H} \nabla_{\theta^h_i}\bs{H} \bs{U}_{H} \bs{\Sigma}_{H})\}\\
    &\quad= \sigma^{-2} \tr(\bs{\Sigma}^2_G)\tr(\nabla_{\theta^h_i}\bs{H})\\
    &\qquad- \sigma^{-2} \diag\{(\sigma^2 \bs{I} + \bs{\Sigma}^2_{G} \otimes \bs{\Sigma}^2_H)^{-1}\}^\top
    \{\diag(\bs{\Sigma}^4_G) \otimes \diag(\bs{\Sigma}_{H} \bs{U}^\ast_{H} \nabla_{\theta^h_i}\bs{H} \bs{U}_{H} \bs{\Sigma}_{H})\}.
\end{align}
Consider $\bs{H} = \bs{W}\bs{W}^\ast$ and the derivative with respect to $w_{dk}$
in the first $\CN(\cdot, \cdot)$ in \eqref{eq:lik}.
In this case, a more computationally inexpensive form is available:
\begin{align}
    &\tr\{(\sigma^2 \bs{I} + \bs{G} \otimes \bs{H})^{-1} (\bs{G} \otimes \nabla_{w_{dk}}\bs{H})\}\\
    &\quad\approx 2\sigma^{-2}\tr(\bs{\Sigma}^2_G) - 2\sigma^{-2} \diag\{(\sigma^2 \bs{I} + \bs{\Sigma}^2_{G} \otimes \bs{\Sigma}^2_H)^{-1}\}^\top \\
    &\qquad\times \{\diag(\bs{\Sigma}^4_G) \otimes [(\bs{U}^\ast_{H} \bs{w}_k) \odot \bs{U}_{H, d:} \odot \diag(\bs{\Sigma}^2_{H})]\}.
\end{align}

For the second term in \eqref{eq:app7},
the following approximation is similarly obtained:
\begin{align}
    &\vect(\bs{Y})^\ast (\sigma^2 \bs{I} + \bs{G} \otimes \bs{H})^{-1} (\bs{G} \otimes \nabla_{\theta^h_i}\bs{H})
        (\sigma^2 \bs{I} + \bs{G} \otimes \bs{H})^{-1}\vect(\bs{Y})\\
    &\quad\approx \sigma^{-4} \{\vect(\bs{Y}) - [(\bs{U}_{G} \bs{\Sigma}_{G}) \otimes (\bs{U}_H \bs{\Sigma}_H)]\vect(\bs{\tilde{Y}})\}^\ast
    [(\bs{U}_G \bs{\Sigma}^2_G \bs{U}^\ast_G) \otimes \nabla_{\theta^h_i}\bs{H}]\\
    &\qquad\times \{\vect(\bs{Y}) - [(\bs{U}_{G} \bs{\Sigma}_{G}) \otimes (\bs{U}_H \bs{\Sigma}_H)]\vect(\bs{\tilde{Y}})\}\\
    &\quad= \sigma^{-4}\vect(\bs{Y})^\ast [(\bs{U}_G \bs{\Sigma}^2_G \bs{U}^\ast_G) \otimes \nabla_{\theta^h_i}\bs{H}] \vect(\bs{Y})\\
    &\qquad - \sigma^{-4} \vect(\bs{\tilde{Y}})^\ast [(\bs{\Sigma}^3_G \bs{U}^\ast_G) \otimes (\bs{\Sigma}_H \bs{U}^\ast_H \nabla_{\theta^h_i}\bs{H})] \vect(\bs{Y}) \\
    &\qquad - \sigma^{-4} \vect(\bs{Y})^\ast [(\bs{U}_H \bs{\Sigma}^3_H ) \otimes (\nabla_{\theta^h_i}\bs{H} \bs{U}_H \bs{\Sigma}_H)] \vect(\bs{\tilde{Y}}) \\
    &\qquad + \sigma^{-4} \vect(\bs{\tilde{Y}})^\ast [\bs{\Sigma}^4_G \otimes (\bs{\Sigma}_H \bs{U}^\ast_H \nabla_{\theta^h_i}\bs{H} \bs{U}_H \bs{\Sigma}_H))] \vect(\bs{\tilde{Y}})\\
    &\quad= \sigma^{-4}\tr(\bs{\Sigma}^2_G\bs{U}^\top_G \bs{Y}^\ast \nabla_{\theta^h_i}\bs{H} \bs{Y} \overline{\bs{U}_G})
    - \sigma^{-4} \tr(\tilde{\bs{Y}} \bs{\Sigma}_H \bs{U}^\ast_H \nabla_{\theta^h_i}\bs{H} \bs{Y} \bs{U}^\top_G \bs{\Sigma}^3_G )\\
    &\qquad - \sigma^{-4} \tr(\bs{\Sigma}^3_G \bs{U}^\top_G \bs{Y}^\ast \nabla_{\theta^h_i}\bs{H} \bs{U}^\ast_H \bs{\Sigma}_H \tilde{\bs{Y}})
    + \sigma^{-4}\tr(\bs{\Sigma}^4_G \tilde{\bs{Y}}^\ast \bs{\Sigma}_H \bs{U}^\ast_H \nabla_{\theta^h_i}\bs{H} \bs{U}_H \bs{\Sigma}_H \tilde{\bs{Y}}^\ast).
\end{align}
When $\bs{H} = \bs{W}\bs{W}^\ast$ and taking derivative with respect to $w_{dk}$,
\begin{align}
    &\vect(\bs{Y})^\ast (\sigma^2 \bs{I} + \bs{G} \otimes \bs{H})^{-1} (\bs{G} \otimes \nabla_{w_{dk}}\bs{H})
        (\sigma^2 \bs{I} + \bs{G} \otimes \bs{H})^{-1}\vect(\bs{Y})\\
    &\quad\approx 2\sigma^{-4} ( \bs{Y}^\top_{d:} \bs{U}_G \bs{\Sigma}^2_G \bs{U}^\ast_G \bs{Y}^\ast \bs{w}_k
    - \bs{Y}^\top_{d:}\bs{U}_G \bs{\Sigma}^3_G \tilde{\bs{Y}}^\ast \bs{\Sigma}_H \bs{U}^\ast_H \bs{w}_k\\
    &\qquad - \bs{U}^\top_{H, d:} \bs{\Sigma}_H \tilde{\bs{Y}} \bs{\Sigma}^3_G \bs{U}^\top_G \bs{Y}^\ast \bs{w}_k
    + \bs{U}^\top_{H, d:} \bs{\Sigma}_H \tilde{\bs{Y}} \bs{\Sigma}^4_G \tilde{\bs{Y}}^\ast \bs{\Sigma}_H \bs{U}^\ast_H \bs{w}_k).
\end{align}

These derivative approximations of $\ell(\bs{\theta}^g, \bs{\theta}^h)$ imply the effectiveness of our low-rank approximation
in terms of computational costs, which are lower than
those of the Stegle method \citep{stegle2011, rakitsch2013}.

\end{document}